\documentclass[10pt,journal,compsoc]{IEEEtran}
\ifCLASSOPTIONcompsoc
\usepackage[nocompress]{cite}
\else
\usepackage{cite}
\fi
\ifCLASSINFOpdf
\else
\fi

\usepackage{times}
\usepackage{epsfig}
\usepackage{graphicx}
\usepackage{amsmath,bm}
\usepackage{amssymb}
\usepackage{booktabs,caption}
\usepackage[flushleft]{threeparttable}

\usepackage{hyperref}
\hypersetup{colorlinks,allcolors=black}

\begin{document}



\title{Detailed Avatar Recovery from Single Image}

%
%
%
%

\author{Hao~Zhu,~\IEEEmembership{Member,~IEEE,}
	Xinxin~Zuo,~\IEEEmembership{Member,~IEEE,}
	Haotian~Yang,
	Sen~Wang,
	Xun~Cao,~\IEEEmembership{Member,~IEEE,}
	and~Ruigang~Yang,~\IEEEmembership{Senior~Member,~IEEE}
	\IEEEcompsocitemizethanks{\IEEEcompsocthanksitem H. Zhu is with Nanjing University, Nanjing, 210023, China; University of Kentucky, Lexington, KY, 40508, US. \protect\\
		E-mail: zhuhaoese@nju.edu.cn
		\IEEEcompsocthanksitem X. Zuo and S. Wang are with the University of Kentucky, Lexington, KY, 40508, US;  Northwestern Polytechnical University, Xi'an, 710072, China; University of Alberta, Edmonton, AB, Canada.\protect\\
		E-mail: xinxinzuo2353@gmail.com, wangsen1312@gmail.com
		\IEEEcompsocthanksitem H. Yang is with Nanjing University, Nanjing, 210023, China. \protect\\
		E-mail: yanght321@gmail.com
        \IEEEcompsocthanksitem X. Cao (corresponding author) is with Nanjing University, Nanjing, 210023, China. \protect\\
		E-mail: caoxun@nju.edu.cn		
		\IEEEcompsocthanksitem R. Yang is with University of Kentucky, Lexington, KY, 40508, USA; Inceptio Technology. \protect\\
		E-mail: ryang2@uky.edu
	}
	\thanks{Manuscript received in 2019/7/6, revised in 2021/1/25, accepted in 2021/7/}}

%
%

\markboth{IEEE Transactions on Pattern Analysis and Machine Intelligence,~Vol.~*, No.~*, Month~year}%
{Shell \MakeLowercase{\textit{et al.}}: Bare Demo of IEEEtran.cls for Computer Society Journals}
%



\IEEEtitleabstractindextext{%
\begin{abstract}
    This paper presents a novel framework to recover \emph{detailed} avatar from a single image. It is a challenging task due to factors such as variations in human shapes, body poses, texture, and viewpoints. Prior methods typically attempt to recover the human body shape using a parametric-based template that lacks the surface details. As such resulting body shape appears to be without clothing. In this paper, we propose a novel learning-based framework that combines the robustness of the parametric model with the flexibility of free-form 3D deformation. We use the deep neural networks to refine the 3D shape in a Hierarchical Mesh Deformation (HMD) framework, utilizing the constraints from body joints, silhouettes, and per-pixel shading information. Our method can restore detailed human body shapes with complete textures beyond skinned models. Experiments demonstrate that our method has outperformed previous state-of-the-art approaches, achieving better accuracy in terms of both 2D IoU number and 3D metric distance.
\end{abstract}
}

\maketitle

\section{Introduction}

\IEEEPARstart{B}UILDING a human avatar from a single image is a challenging problem and has drawn much attention in recent years. A large number of approaches~\cite{ICCV2009Guan, ECCV2016Bogo, 3DV2016Dibra, BMVC2017Tan, CVPR2017Lassner, NIPS2017Tung, CVPR2018Pavlakos, CVPR2018Kanazawa, 3DV2018Omran, Kolotouros_2019_CVPR, Kanazawa_2019_CVPR, Alldieck_2019_CVPR} have been proposed in which the human body shapes are reconstructed by predicting the parameters of a statistical skinned model, such as SMPL~\cite{TOG2015Loper} and SCAPE~\cite{TOG2005Anguelov}. The parametric shape is of low-fidelity, and unable to capture clothing details. Though several works\cite{Alldieck_2019_CVPR, 3DV2016Dibra, CVPR2018Pavlakos} attempts to recover more details than a skinned model, they did not go far on the issue of detailed geometry recovery.
Another collection of methods\cite{ECCV2018Varol, BMVC2018Venkat} estimate volumetric human shape directly from the image using neural networks, while the resulting volumetric representation is fairly coarse and does not contain shape details.

\begin{figure}[t]
\begin{center}
   \includegraphics[width=1.0\linewidth]{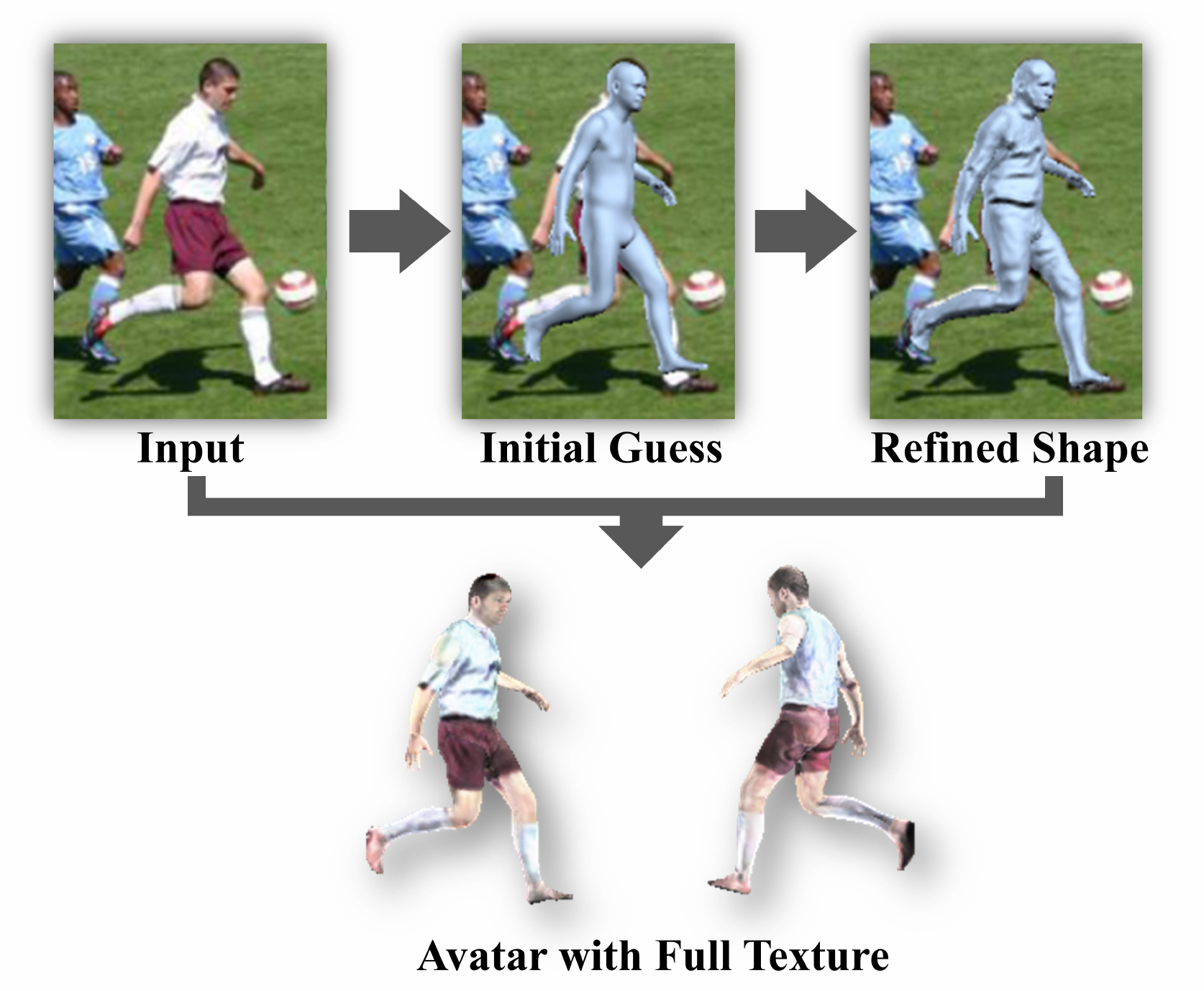}
\end{center}
   \vspace{-0.15in}
   \caption{Our method takes a single image of a person in the wild as input and predicts detailed human body shape with texture, namely human avatar.  Our method recovers body shapes with surface details that fit the input image well, and also hallucinates the complete texture from the visible region.}
\label{fig:tile}
\end{figure}

The limited performance of previous methods is caused by the large variations of the human shape and pose. Parametric or volumetric shapes are not expressive enough to represent the inherent complexity. Besides, most of those previous methods focus on shape recovery while neglecting surface texture or appearance which is another important aspect to build a human avatar.

In this paper, we propose a novel framework to reconstruct \emph{detailed} human avatar from a single image. The key idea is to combine the robustness of the parametric model with the flexibility of free-form deformation. In short, we build on top of the SMPL model to obtain an initial parametric mesh model and perform non-rigid 3D deformation on the mesh to refine the surface shape.
We design a coarse-to-fine refinement scheme in which a deep neural network is used in each stage to estimate the 3D mesh vertex movement by minimizing its 2D projection error in the image space. 
We feed window-cropped images instead of the full image to the network, which leads to a more accurate and robust prediction of deformation.
In addition, we integrate a photometric term to allow high-frequency details to be recovered. 
These techniques combined lead to a method that significantly improves, both visually and quantitatively, the recovered human shape from a single image as shown in Figure \ref{fig:tile}. 
Finally, we regress the complete texture of the reconstructed human model from the input image with our proposed texture synthesis network.

Different from our previous version ~\cite{zhu2019detailed}, we extend our system by texturing the reconstructed 3D human body and recovering its complete appearance from a single image. This is the so-called human avatar which includes both the human shape and its texture. The major issue for the texture recovery problem is that for the predicted topologically-uniformed human mesh, only less than half of the texture can be seen from the single input image, and the visible textures are also semantically misaligned due to the deviation between the predicted mesh and the image. Therefore, synthesizing complete and realistic textures is quite challenging in this case. In this paper, combining the advantages of both flow-based warping networks and image generation networks, we design a novel framework to synthesize the missing texture with high tolerance to image misalignment and background interference.

The contributions of this paper include:
\begin{itemize}
\item 
We develop a novel \textit{project - predict - deform} strategy to predict the deformation of the 3D mesh model by using 2D features. 

\item 
We carefully design a hierarchical update structure, incorporating body joints, silhouettes, and photometric-stereo to improve shape accuracy without losing robustness. 

\item 
We propose a texture synthesis network to restore the complete texture from the input image, which is robust to the misalignment caused by the shape reconstruction phase. 

\item 
As demonstrated throughout our experiments, the additional free deformation of the initial parametric model leads to quantitatively more accurate shapes with good generalization capabilities to images in the wild. The texture synthesis network generates the complete texture for the recovered shape, after which we will get the final human avatar.

\end{itemize}

\begin{figure*}[t]
\begin{center}
   \includegraphics[width=1.0\linewidth]{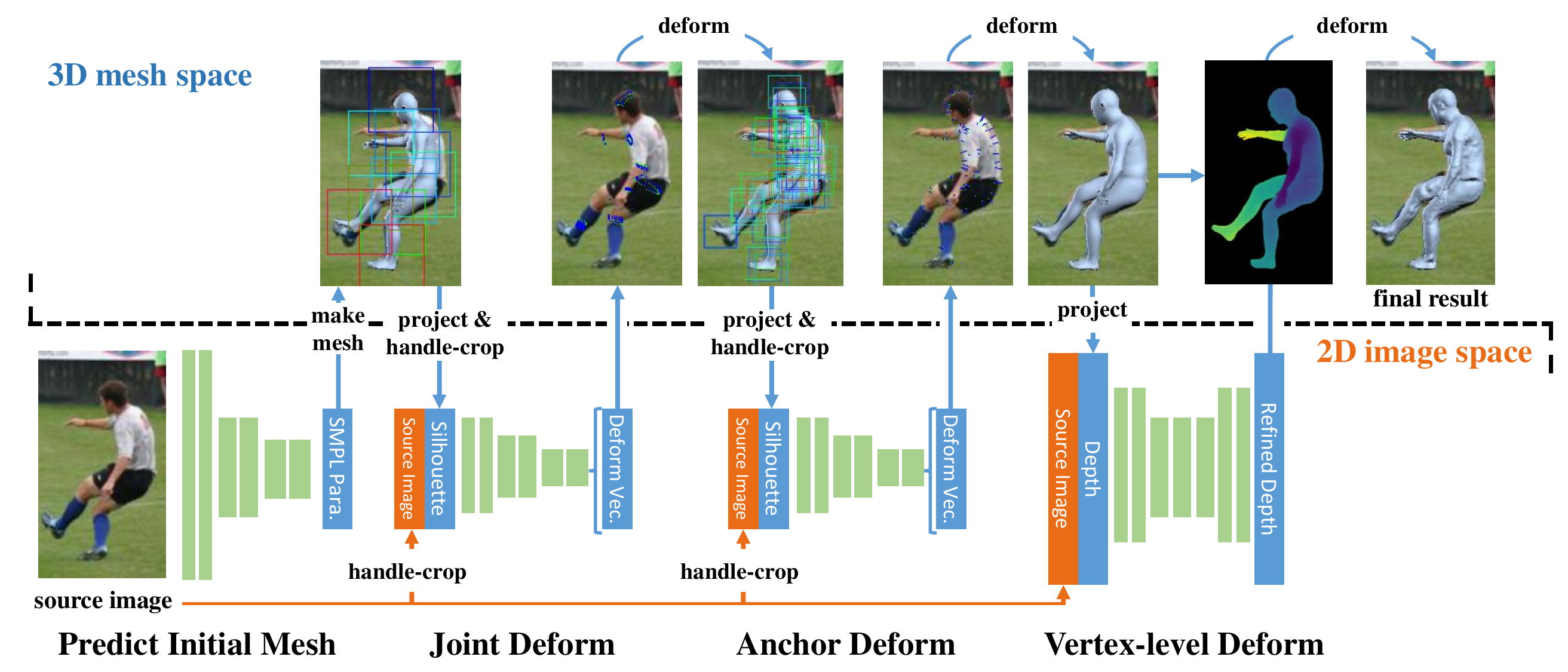}
\end{center}
   \caption{The flow of our shape-recovering method goes from the bottom left to the top right.  The mesh deformation architecture consists of three levels: joint, anchor and per-vertex.  In each level, the 3D mesh is projected to 2D space and sent together with the source image to the prediction network. The 3D mesh gets deformed by the predicted results to produce refined human body shapes.}
\label{fig:pipeline}
\end{figure*}

\section{Related Work}
Previous approaches can be divided into two categories based on the way the human body is represented: parametric methods and non-parametric methods. 

As for parametric methods, they rely on a pre-trained generative human model, such as the SCAPE~\cite{TOG2005Anguelov} or SMPL~\cite{TOG2015Loper} models. The goal is to predict the parameters of the generative model. The SCAPE model has been adopted by Guan \textit{et al.}~\cite{ICCV2009Guan} to recover the human shape and poses from the monocular image as provided with some manually clicked key points and the constraint of smooth shading.  Instead of relying on manual intervention, Dibra \textit{et al.} \cite{3DV2016Dibra} have trained a convolutional neural network to predict SCAPE parameters from a single silhouette image. Similar to the SCAPE model, Hasler \textit{et al.} \cite{CVPR2010Hasler} have proposed a multi-linear model of human pose and body shape which was generated by factorizing the measurements into the pose and shape dependent components. The SMPL model~\cite{TOG2015Loper} has recently drawn much attention due to its flexibility and efficiency. For example, Bogo \textit{et al.}~\cite{ECCV2016Bogo} have presented an automatic approach called SMPLify which fits the SMPL model by minimizing an objective function that penalizes the error between the projected model joints and detected 2D joints obtained from a CNN-based method together with some priors over the pose and shape. Building upon this SMPLify method, Lassner \textit{et al.}~\cite{CVPR2017Lassner} have formed an initial dataset of 3D body fitting with rich annotations consisting of 91 key points and 31 segments. Using this dataset, they have shown improved performance on part segmentation, pose estimation, and 3D fitting. Tan \textit{et al.}~\cite{BMVC2017Tan} have proposed an indirect learning procedure by first training a decoder to predict body silhouettes from SMPL parameters and then using pairs of real images and ground truth silhouettes to train a full encoder-decoder network to predict SMPL parameters at the information bottleneck. Pavlakos \textit{et al.}~\cite{CVPR2018Pavlakos} have separated the SMPL parameters prediction network into two sub-networks. Taking the 2D image as input, the first network was designed to predict the silhouette and 2D joints, from which the shape and pose parameters were estimated respectively. The latter network combined the shape and 2D joints to predict the final mesh. Kanazawa \textit{et al.}~\cite{CVPR2018Kanazawa} have proposed an end-to-end framework to recover the human body shape and pose in the form of SMPL model using only 2D joints annotations with an adversarial loss to effectively constrain the pose. Instead of using joints or silhouettes, Omran \textit{et al.}~\cite{3DV2018Omran} believed that a reliable bottom-up semantic body part segmentation was more effective for shape and pose prediction. Therefore, they predicted a part segmentation from the input image in the first stage and took this segmentation to predict SMPL parameter of the body mesh.  Alldieck \textit{et al.}~\cite{Alldieck_2019_CVPR} have used a two-stage strategy to recover the human shape. Firstly, they used a part-based model to regress the 3D parametric human model; In the second stage, an iterative refinement was applied using the unwrapped texture to reconstruct the body shape.  Kanazawa \textit{et al.}~\cite{Kanazawa_2019_CVPR} have proposed a framework that could learn a representation of 3D humans dynamics from a video via the temporal encoding of image features, which has taken advantage of the temporal information to enhance the recovered shape quality.
Alldieck \textit{et al.}~\cite{Alldieck_2019_CVPR} have presented a learning-based approach to estimate body shapes including hair and clothing and they also took a monocular video as input.
Yu \textit{et al.}~\cite{yu2019simulcap} have proposed a multi-layer representation of garments and body to capture human performance using an RGBD camera. The physics-based cloth simulation was incorporated into the performance capture pipeline, to simulate plausible cloth dynamics and cloth-body interactions.
Bhatnagar \textit{et al.}~\cite{bhatnagar2019multi} have proposed to predict body shape and clothing, layered on top of the SMPL model from a single or a few frames. The predicted garment geometry is related to the body shape and can be transferred to new body shapes and poses.
Alldieck \textit{et al.}~\cite{alldieck2019tex2shape} have proposed to reconstruct the human shape from a single image in UV space.  In their framework, a UV transformer firstly transformed the input image to the partially visible UV texture, and then a PatchGAN\cite{isola2017image} was used to synthesize the complete normal and displacement map that can be applied to the SMPL model.
Mir \textit{et al.}~\cite{mir2020learning} have focused on transferring textures of clothing images to 3D garments worn on top SMPL, which enables 3D virtual try-on in real-time.

Non-parametric methods directly predict the shape representation from the image. 
Some researchers have used depth maps as a more general and direct representation of shapes. For example, Varol \textit{et al.}~\cite{CVPR2017Varol} have trained a convolutional neural network by building up a synthetic dataset of rendered SMPL models to predict the human shape in the form of depth image and body part segmentation.  G{\"u}ler \textit{et al.}~\cite{CVPR2017Guler, CVPR2018Guler} have treated the shape prediction problem as a correspondence regression problem, which would produce a dense 2D-to-3D surface correspondence field for the human body. Another way of representing 3D shapes is to embed the 3D mesh into a volumetric space~\cite{ECCV2018Varol,BMVC2018Venkat}. For example, Varol \textit{et al.}~\cite{ECCV2018Varol} have restored volumetric body shape directly from a single image.
The mesh model was extracted from the predicted volumetric model and fitted to an SMPL model as a post-processing procedure. 
Similarly, Venkat \textit{et al.} \cite{BMVC2018Venkat} have recovered the volumetric grid of the human body from a single image and they have put much effort into texture view synthesis to get a textured 3D model. However, they only performed the test on images captured in the lab environment rather than images in the wild. 
In addition to depth and volumetric representations, Kulkarni \textit{et al.}~\cite{CVPR2015Kulkarni} have proposed a probabilistic programming language that can express generative models for arbitrary 2D/3D objects. The language was used to predict pose as well as shape from simple images while the accuracy was not evaluated. 
Dibra \textit{et al.}~\cite{CVPR2017Dibra} have proposed to learn a mapping from silhouettes to an embedding space from which 3D human body mesh will get restored. 
Their method focuses more on robust body measurements rather than shape details or poses.
Kolotouros \textit{et al.}~\cite{Kolotouros_2019_CVPR} have proposed to relax the reliance on the model’s parameter space, and directly regressed the 3D location of the mesh vertices instead of predicting the model's parameters using a Graph-CNN.  Natsume \textit{et al.}~\cite{Natsume_2019_CVPR} have introduced the implicit representation that uses 2D silhouettes and 3D joints of a body pose to describe the immense shape complexity and variations of clothed people.  They used the deep visual hull algorithm to predict 3D shape from the synthesized silhouettes which are consistent with the input segmentation, and also inferred the texture of the back view using a conditional generative adversarial network.
Lazova \textit{et al.}~\cite{lazova2019360} have proposed to firstly predict the dense correspondence and garment segmentation, then a neural network was designed to predict the completed texture and displacement maps respectively. The displacement map and the complete texture were then merged on the base of the SMPL model to form the fully-textured 3D avatar.
Habermann \textit{et al.}~\cite{habermann2019livecap} have proposed a real-time human performance capture approach that reconstructed dense, space-time coherent deforming geometry of clothed people from a single monocular RGB stream. Compared with our proposed method, this method requires a pre-reconstructed model as a reference.  
Smith \textit{et al.}~\cite{smith2019facsimile} have proposed to use an image-translation network to recover the 3D geometry of a human body, and adopted per-pixel surface normals instead of per-pixel depth for training loss, which has made it possible to estimate detailed body geometry.
Zheng \textit{et al.}~\cite{zheng2019deephuman} have proposed to use the single image together with dense semantic representation generated from SMPL mesh as input, and used a 3D convolutional network to predict the volumetric shape of the target human.
Tang \textit{et al.}~\cite{tang2019neural} have proposed to train a network to predict the depth map of the human body, and Tan \textit{et al.}~\cite{tan2020self} further proposed to train the network to predict human depth map in a self-supervised manner. The photometric loss within several frames was used to supervise the regression of the depth map after the human motion was compensated.

In recent years, implicit functions have also shown great potential in single view reconstruction of the human body. Saito \textit{et al.}~\cite{saito2019pifu} have proposed a framework with a pixel-aligned implicit function to estimate the shape of the clothed human shape. 
As a follow-up, a multi-level architecture has been introduced to reveal the surface details~\cite{saito2020pifuhd}, and it has been optimized for real-time performance capture~\cite{li2020monocular}. Similarly, Huang \textit{et al.}~\cite{huang2020arch} have proposed to firstly estimate the correspondence between the input image and the canonical model, and then reconstructed the human shape using the implicit function in the canonical space. PaMIR~\cite{zheng2021pamir} proposed to combine the parametric body model with the free-form deep implicit function, which improves the generalization ability of humans with challenging poses. Similarly, both PaMIR and our method utilize SMPL based parametric models. However, PaMIR predicts the implicit function to represent the cloth, while our method expresses the surface details with a displacement map of the parametric model and recovers the clothed human shape through a free-form deformation, thereby maintaining the mesh topology and the ability of rigging.

While significant progress has been made in this very difficult problem, the recovered human shape is still lacking in accuracy and details. In contrast to all the above methods, we present a method to predict human body shape from course to fine at multiple scales, and we also propose a texture completion method to generate the human avatar with complete texture.

\section{Hierarchical Deformation Framework}
In this section, we will present our hierarchical deformation framework to recover \emph{detailed} human body shapes by refining a template model in a coarse-to-fine manner. As shown in Figure~\ref{fig:pipeline}, there are four stages in our framework: First, an initial SMPL mesh is estimated from the source image. Starting from this, the next three stages serve as refinement phases which predict the deformation of the mesh so as to produce a detailed human shape. We have used the HMR method~\cite{CVPR2018Kanazawa} to predict the initial human mesh model, which has demonstrated state-of-the-art performance on human shape recovery from a single image. However, like other human shape recovery methods~\cite{CVPR2018Pavlakos, 3DV2018Omran, ECCV2016Bogo} that utilize the SMPL model, the HMR method predicts the shape and pose parameters to generate a skinned mesh model with limited flexibility to closely fit the input image or express surface details. For example, the HMR often predicts deflected joint position of limbs when the human pose is unusual. Therefore, we have designed our framework to refine both the human shape and the pose. 

The refining stages are arranged hierarchically from coarse to fine. We define three levels of key points on the mesh, referred to as \emph{handles} in this paper. We will describe exactly how we define these handles in the next section. In each level, we design a deep neural network to refine the 3D mesh geometry using these handles as control points. We train the three refinement networks separately and successively to predict the residual deformation based on its previous phase.

To realize the overall refinement procedure, a challenging problem is how to deform the 3D human mesh from handles in 2D space using deep neural networks. We address this using Laplacian mesh deformation. In detail, the motion vector for each handle is predicted from the network driven by the joints and silhouettes of the 2D image. Then the human mesh will get deformed with the Laplacian deformation approach given the movements of the handles while maintaining the local geometry of the human model. The deforming strategy has been used in multi-view shape reconstruction problem~\cite{CVPR2018Alldieck, TCSVT2017Zhu, liao2009modeling, ICCV2001Plankers, zhu2017role, ECCV2016Rhodin, 3DV2016Robertini}, while we are the first to predict the deformation from a single image with the deep neural network.  

\subsection{Handle Definitions}
In this section, we will describe the handles that we have used in each level. They could be predefined in the template model thanks to the uniform topology of SMPL mesh model.

\begin{figure}[t]
\begin{center}
   \includegraphics[width=1.0\linewidth]{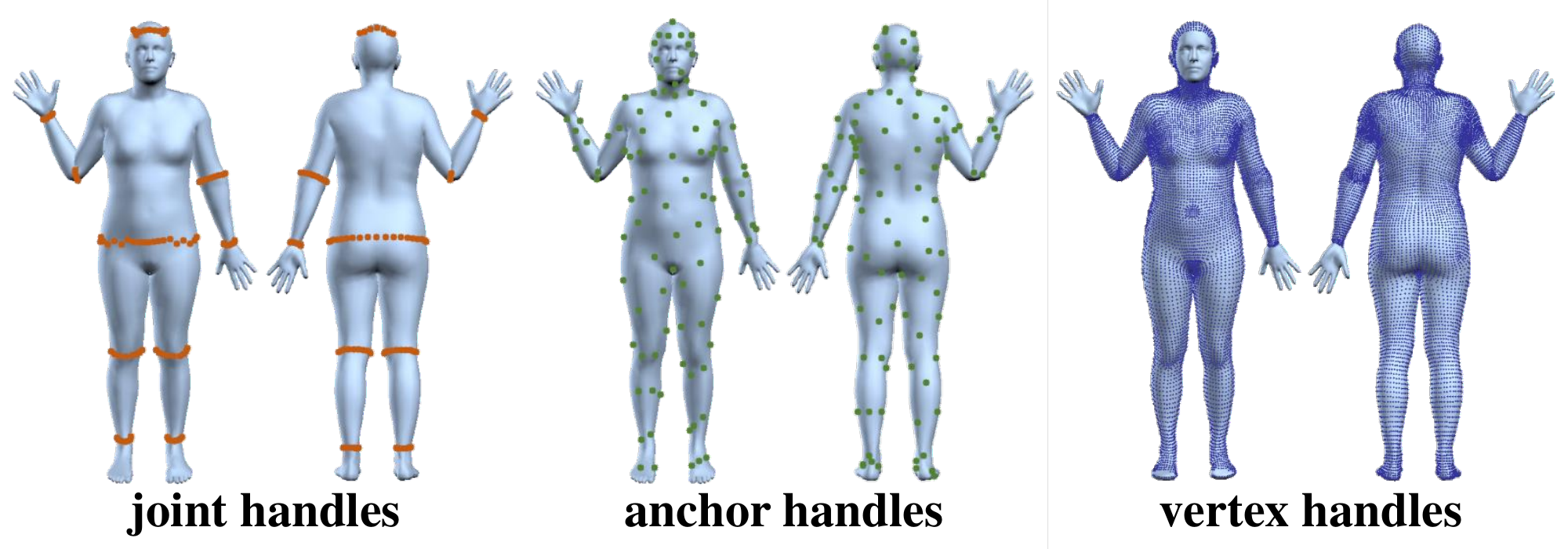}
\end{center}
   \caption{The handles definition in different levels for mesh deformation.  
   }
\label{fig:handles}
\end{figure}

\textbf{Joint handles.}  We select 10 joints as the control points -- head, waist, left/right shoulders, left/right elbows, left/right knees, and left/right ankles. The vertices around the joints under the T-pose SMPL mesh are selected as handles, as shown in Figure~\ref{fig:handles}. We take the geometric center of each set of handles as the position of its corresponding body joint. The motion of each joint handle is represented as a 2D vector, which refers to the vector from the joint position of projected mesh to ground truth joint position on the image plane.

\textbf{Anchor handles.}  We select 200 vertices on the human template mesh under T-pose as anchor handles. To select the anchors evenly over the template, we firstly build a vector set $C = \{v_1, v_2, ......, v_n\}$ with $v_i$ concatenated by the position and surface normal of the vertex $i$ and $n$ is the number of SMPL model vertices. Then K-means is applied to cluster set $C$ into 200 classes. Finally, we set the closest vertex to the center of each cluster as the anchor handles. Besides, we have removed the vertices in the face, fingers, toes from the T-pose SMPL model to prevent the interference of high-frequency shape. To be noticed that, for each anchor, it is only allowed to move along the surface normal direction, so we just need to predict a single value indicating the movement of the anchor point along the normal direction.

\textbf{Vertex handles.} The vertices in the SMPL mesh are too sparse to apply pixel-level deform, so we subdivide each face of the mesh into 4 faces.  The subdivision increases the number of vertices of the template mesh to 27554, and all these vertices are regarded as handles.

\subsection{Joint and Anchor Prediction}
\label{sec:deformation}

\textbf{Network.} Both joint and anchor prediction networks use the VGG~\cite{arXiv2014VGG} structure which consists of a feature extractor and a regressor. The network takes the mesh-projected silhouette and source image as input, which are cropped into patches as centered with our predefined handles. Basically, as joint and anchor handles are pre-defined on the topologically uniformed SMPL mesh, we can get the 3D position of the joint/anchor handles by indexing on the mesh.  Then the 2D position of the handles can be obtained by projecting 3D handle points, and the cropping windows are centered on these points with the predefined square size.  For a $224\times224$ input image, the image is cropped into patches with the size of $64\times64$ for joint prediction, and $32\times32$ for anchor prediction. Compared with the full image or silhouette input, the handle cropped input allows the network to focus on the region of interest. 
We will demonstrate the effectiveness of the cropped input in Section \ref{sec:stage_exp}.  

\textbf{Loss.}  The output of the joint net is a 2D vector representing the joint motion in the image plane.  L2 loss is exploited to train the joint net with the loss function formulated as:

\begin{equation}
L_{joint} = ||\bm{p} - \bm{\widehat{p}}||_2
\label{equ:loss_func}
\end{equation}

\noindent where $\bm{p}$ is the predicted motion vector from the network and $\widehat{\bm{p}}$ is the displacement vector from the mesh-projected joint position to its corresponding ground-truth joint. Both vectors are 2-dimensional.

For the anchor net, our immediate goal is to minimize the area of the mismatched part between the projected silhouette and the ground truth silhouette. One strategy is to follow Alldieck \textit{et al.}'s work~\cite{CVPR2018Alldieck}, where the mesh is optimized to fit the multi-view silhouettes. However, in the single-view silhouette fitting problem, it requires strong constraints to maintain the global structure stable. Therefore, we take advantage of the Laplacian deforming strategy, and use sparse-sampled anchor handles as control points. 
We use a transformation vector to represent the deformation of the silhouette.  The transformation vector is in the same direction as the normal vector of the anchor vertex, and its length is calculated as the distance from the predicted silhouette to the ground-truth silhouette along the vertex normal direction. The transformation vector is regarded as the movement of the anchor, and L2 loss is used to train the network.  In the Laplacian deformation stage, two kinds of anchor handles do not participate in deformation as control points: One kind is the point with deformation distance $>0.1m$, which is considered as the internal anchor handle; The other kind is the point that is too close to the edge of the silhouette.  Since the Laplacian deforming will keep the local geometry as much as possible, the overall shape would be deformed equably.

Besides, instead of using the RGB image as input, the joint and anchor prediction network could also take the ground-truth silhouette of the human figure as input if available.  The silhouette provides more explicit information for the human figure, which prevents the network from getting confused by the cluttered background environment. We will demonstrate its effectiveness on joint and anchor deformation prediction in the experiment section. In this paper, we consider the RGB-only as input by default, and use `+Sil.' to indicate the case where the additional silhouette is used.

\subsection{Vertex-level Deformation}

To add high-frequency details to the reconstructed human models, we exploit the shading information contained in the input image. First, we project the current 3D human model into the image space, from which we will get the coarse depth map. We then train a \emph{shading net} that takes the color image and current depth map as input and predicts a refined depth map with surface details. 
We have built up a relatively small dataset that contains color images, over-smoothed depth maps, and corresponding ground truth depth maps that have good surface details. More detailed explanations on this dataset could be found in Section \ref{ssec:Imp}.
We adopt a multi-stage training scheme with limited labeled data. 

Following the training scheme proposed in~\cite{CVPR2018sfs}, we firstly train a simple U-Net based encoder-decoder network~\cite{Unet2015} on our captured depth dataset taking the over-smoothed depth map and its corresponding color image as input. The network is trained as supervised by the ground truth depth maps. Next, we adopt this network on the real images of our human body dataset to obtain enhanced depth maps. As we only have limited supervised data, the network may not generalize well to our real images. Therefore, to finally get depth maps with great surface details consistent with the color images, we train our \emph{shading net}, which is also a U-Net based network on real images. The network is trained with both the supervision loss using the depth maps output by the first U-Net and also a photometric reconstruction loss~\cite{ECCV2018DDRNet} that aims to minimize the error between the original input image and the reconstructed image. The per-pixel photometric loss $L_{photo}$ is formulated as below:

\begin{equation}
\label{Eq:lighting}
L_{photo} =||\rho \sum_{k=1}^{9} l_{k}H_{k}(\bm{n}) -I ||_2
\end{equation}

\noindent where $\rho$ is the albedo computed by a traditional intrinsic decomposition method~\cite{bell2014intrinsic}. Similar to~\cite{ramamoorthi2001efficient, zuo2017detailed}, we use the second spherical harmonics (SH) for illumination representation under the Lambertian surface assumption. $H_{k}$ represents the basis of spherical harmonics. $l_1,l_2...l_9$ denote the SH coefficients, which are computed under a least square minimization as: 

\begin{equation}  
\label{Eq:lighting2}
\bm{l}^{*} = \mathop{\arg\min}_{\bm{l}}  ||\rho \sum_{k=1}^{9} l_{k}H_{k}(\bm{n}_{coarse}) -I ||_2^2
\end{equation}

We use the coarse depth map rendered from the currently recovered 3D model to compute the surface normal $\bm{n}_{coarse}$. 

\noindent \textbf{Enhance by 3D-supervision.}  In the above settings, we use a small number of depth maps captured by Kinect, then adopt photometric loss to achieve semi-supervised training. In recent years, a number of high-quality 3D human body commercial datasets\cite{twindom, renderpeople} have emerged, which provides higher quality and a larger quantity of 3D human models. We try to use these high-quality data to enhance the prediction effect of shading net, referred to as 3D-supervised training.  Specifically, we use 500 3D models from the Twindom dataset\cite{twindom}, and render each 3D model with Lambertian diffuse shading with surface normal and spherical harmonics\cite{CVPR2017Varol, Natsume_2019_CVPR} in 180 different light conditions and viewpoints. $80\%$ of the data are used for training and the other are used for testing.  Same as the semi-supervised training, the shading net in the 3D-supervised training takes the projected depth and the source image as input, and predicts a refined depth map with surface details.  L1 loss between the predicted depth and the ground-truth rendered depth is used in the training of the shading net.  The predicted depth is then used to refine the mesh by deforming the vertices accordingly.

We find that the 3D-supervised training results in a certain degree of improvement, which will be discussed in Section \ref{sec:3dspv}.  Comparing to the semi-supervised scheme, the 3D-supervised scheme enhances the performance, but requires the expensive commercial 3D dataset. Users may choose the training scheme by trading off the cost and the quality requirements.

\subsection{Implementation Details}\label{ssec:Imp}
We use the pre-trained model in the HMR-net, then train Joint-Net, Anchor-Net, and shading net successively.  We use the `Adam' optimizer to train these networks, with the learning rate as $0.0001$.  The handle weight in Laplacian edit is $10$ for joint deforming and is $1$ for anchor deforming.

To provide better training data to the shading net, we have captured a small depth dataset with a Kinect V2 sensor. It consists of $2200$ depth frames with three human subjects wearing different clothes under various poses. The captured depth maps are further enhanced using traditional shading refinement techniques~\cite{or2015rgbd, zollhofer2014real} to recover small surface details, which can be taken as ground truth depth maps for supervised learning. We have magnified the shape details by $10$ times during the test time.

\section{Texture Completion}
Synthesizing complete texture for the reconstructed human model from a single image is also a challenging problem, since only less than half of the texture is visible and can be retrieved from the input image. Besides, the imperfect fitting between the reconstructed human shape and the input image brings more difficulties to this problem.

\begin{figure}[t]
\begin{center}
   \includegraphics[width=1.0\linewidth]{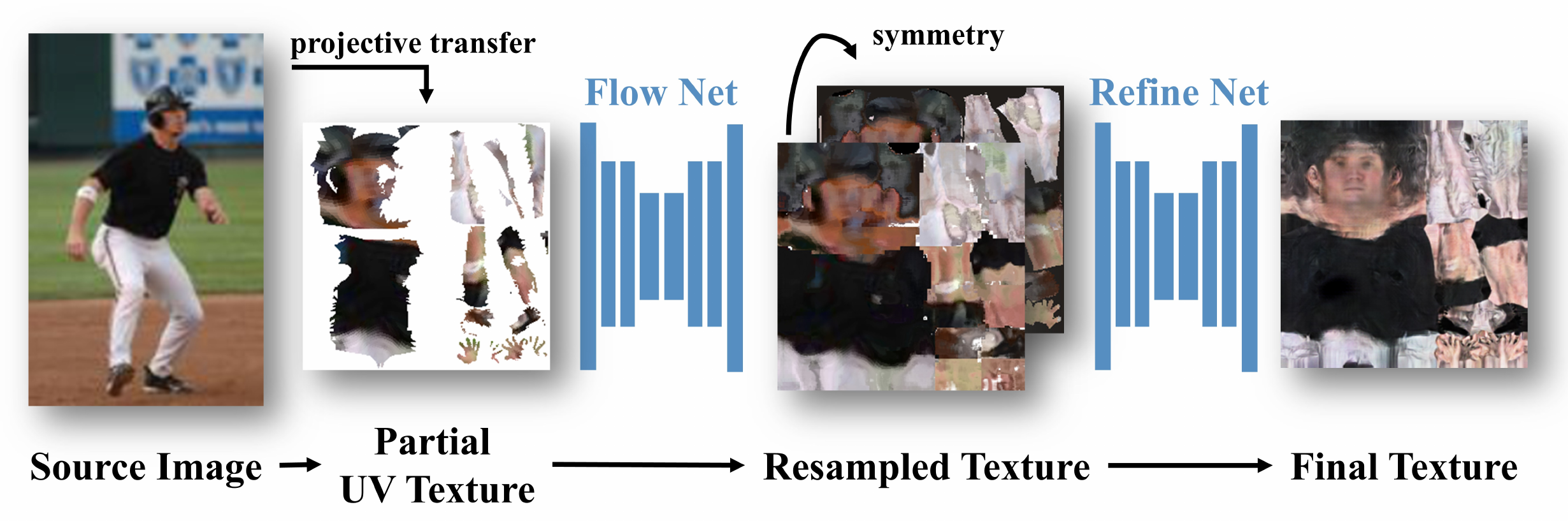}
\end{center}
   \caption{The pipeline for texture completion.}
\label{fig:tex_cplt}
\end{figure}

As shown in Figure \ref{fig:tex_cplt}, our main idea is to take advantage of the flow prediction and image generation networks to synthesize a complete texture map from the visible part. Previous methods\cite{zhou2016view, park2017transformation, CVPR2018Zhu} have demonstrated that the neural networks can be used to synthesize novel views for rigid objects from specific categories, like cars and chairs.  However, these methods failed for the synthesis of humans. The reason is that as compared to those rigid objects, very complex body poses are involved for the humans which result in misalignment for the features from different samples and significantly increase the difficulty for the synthesis~\cite{CVPR2018Zhu}. Therefore, in this paper, we propose to predict the complete texture in the UV coordinates to avoid the interference of the misaligned features. As the reconstructed meshes are in uniform topology, the texture features are roughly aligned in the UV coordinates. The minor misalignment caused by inaccurate shape recovering can be corrected by the deep neural networks.

\subsection{Training Data}\label{ssec:Texture_trainging_data}
We use the synthetic models from SURREAL dataset~\cite{CVPR2017Varol} to train our texture synthesizer. Specifically, we use $929$ models with different appearances in SURREAL dataset to generate train/test data.  For each appearance, we construct $20$ SMPL models with randomly generated shape parameters and also with various poses randomly selected from the UP dataset~\cite{CVPR2017Lassner}. We render each generated model with $5$ views randomly chosen from $54$ viewpoints corresponding to $3$ pitch angles ($-20^{\circ}$, $0^{\circ}$, $20^{\circ}$) and $18$ azimuth angles in the range [$0^{\circ}$, $340^{\circ}$] with interval $20^{\circ}$.  The image background is randomly sampled from the Places dataset~\cite{TPAMI2017Zhou}. 
We have generated $92900$ images in total using the above-mentioned process. The generated data is divided into the train and validate set, among which the train set accounts for $90\%$, and the validate set accounts for $10\%$. We train the network with the train set and evaluate the performance of the trained model using validate set, and finally, show the results on the real image from the WILD dataset in the experiment section.

Since our reconstructed mesh can not be perfectly aligned with the input image, the color spilling artifacts exists in the visible parts.  To make our network robust to the misalignment, we augment the training data by introducing random bias.  Specifically, we perturb the generated model and viewpoint parameters with uniformly distributed noise proportional to the parameter value. Then we detect visible faces on the mesh based on the perturbed parameters. For each visible face, we project corresponding pixels in the image to the UV space to get the partial texture and binary visibility mask. As demonstrated in the experiments, our texture synthesis network has better generalization ability when tested on wild images with this strategy.

\subsection{Appearance Flow for Texture Completion}
Our goal is to generate a complete and plausible texture map from this partial texture map.  To this end, we map the visible texture into the UV coordinates. Different from other image inpainting problems where the masked region is only a small proportion concerning the original image, in this case, the invisible part is typically more than half and irregular.  It has been shown that convolutional neural networks process image features with local convolution kernel layer by layer and thus are not effective for borrowing features from distant spatial location~\cite{yu2018generative}.  Also, directly inpainting images with large missing parts tend to produce artifacts such as blurriness and color discrepancy.  Some methods such as contextual attention~\cite{yu2018generative} and shift operation~\cite{yan2018shift} have been proposed to deal with this problem.  However, these approaches are designed for rectangular masks and are not trivial to generalize to masks with arbitrary shapes. 

Inspired by recent advances in novel view synthesis~\cite{zhou2016view, park2017transformation} and human pose transfer~\cite{neverova2018dense, li2019dense}, we propose to use appearance flow to complete the texture map.  Specifically, taking the partial texture $I_{part}$ and binary mask $M$ as input, our network predicts a dense flow field of the same size as the texture.  Then we use the differentiable bilinear sampling layer introduced in~\cite{jaderberg2015spatial} to get the pixel value in the output image $I_{flow}$ based on the predicted flow field.  L1 loss is used to train the flow net:

\begin{equation}
\label{Eq:lighting3}
L_{flow} =||I_{gt} - I_{flow}||_1
\end{equation}

\noindent where $I_{gt}$ is the ground truth complete texture in UV coordinates.

\subsection{Symmetry Aware Texture Refinement}

As our flow net is unable to hallucinate the missing texture, some pixels cannot be filled properly with existing pixels.  To address this problem, we use a refinement network to further improve the texture predicted by the flow net. Considering that the human body is usually symmetrical, we flip the texture according to the symmetric correspondence and concatenate it to the original texture and mask, which is the input to our refinement network.  We find that this strategy gives our texture a certain degree of symmetry, especially in the areas where the left and right seams are significantly improved.  The refinement network generates a complete texture $I_{final}$ of the same size as $I_{flow}$.  Following the state-of-the-art image enhancement~\cite{sajjadi2017enhancenet} and image inpainting methods~\cite{nazeri2019edgeconnect}, we use a joint loss which consists of L1 loss $L_{l_1}$, adversarial loss $L_{adv}$, perceptual loss $L_{perc}$~\cite{johnson2016perceptual, dosovitskiy2016generating}, and style loss $L_{style}$~\cite{sajjadi2017enhancenet}. Our full objective function is formulated as:

\begin{equation}
\begin{aligned}
\mathop{\min}_{G} &(\lambda_{adv} (\mathop{\max}_{D} L_{adv}(G,D)) + \lambda_{perc} L_{perc}(G) \\ &+  \lambda_{style} L_{style}(G) + L_{l_1}(G))
\end{aligned}
\end{equation}

\noindent where $G$ is the refinement net, D is the discriminator. $\lambda_{adv}$, $\lambda_{perc}$ and $\lambda_{style}$ are the weights of adversarial loss, perceptual loss, and style loss, respectively.

\subsection{Implementation Details}
\textbf{Network architecture.}  We use the network proposed by Nazeri \textit{et al.} ~\cite{nazeri2019edgeconnect} as the backbone of both flow net and refinement net.  We use a $70\times70$  PatchGAN architecture~\cite{isola2017image} for our discriminator.  Spectral Normalization~\cite{miyato2018spectral} is used in the refinement net and discriminator to stabilize the training.

\textbf{Training setup.}  Our flow net and refinement net are trained successively.  The loss weights are set to $\lambda_{adv} = 0.1$, $\lambda_{perc} = 0.1$ and $\lambda_{style} = 250$ in all experiments. We train our networks using $256\times256$ images with batches of size $10$.  `Adam' optimizer is used and learning rate is set to $0.0001$.  We randomly jitter the image color for data augmentation.
\section{Experiment}

\begin{figure*}[t]
\begin{center}
   \includegraphics[width=1.0\linewidth]{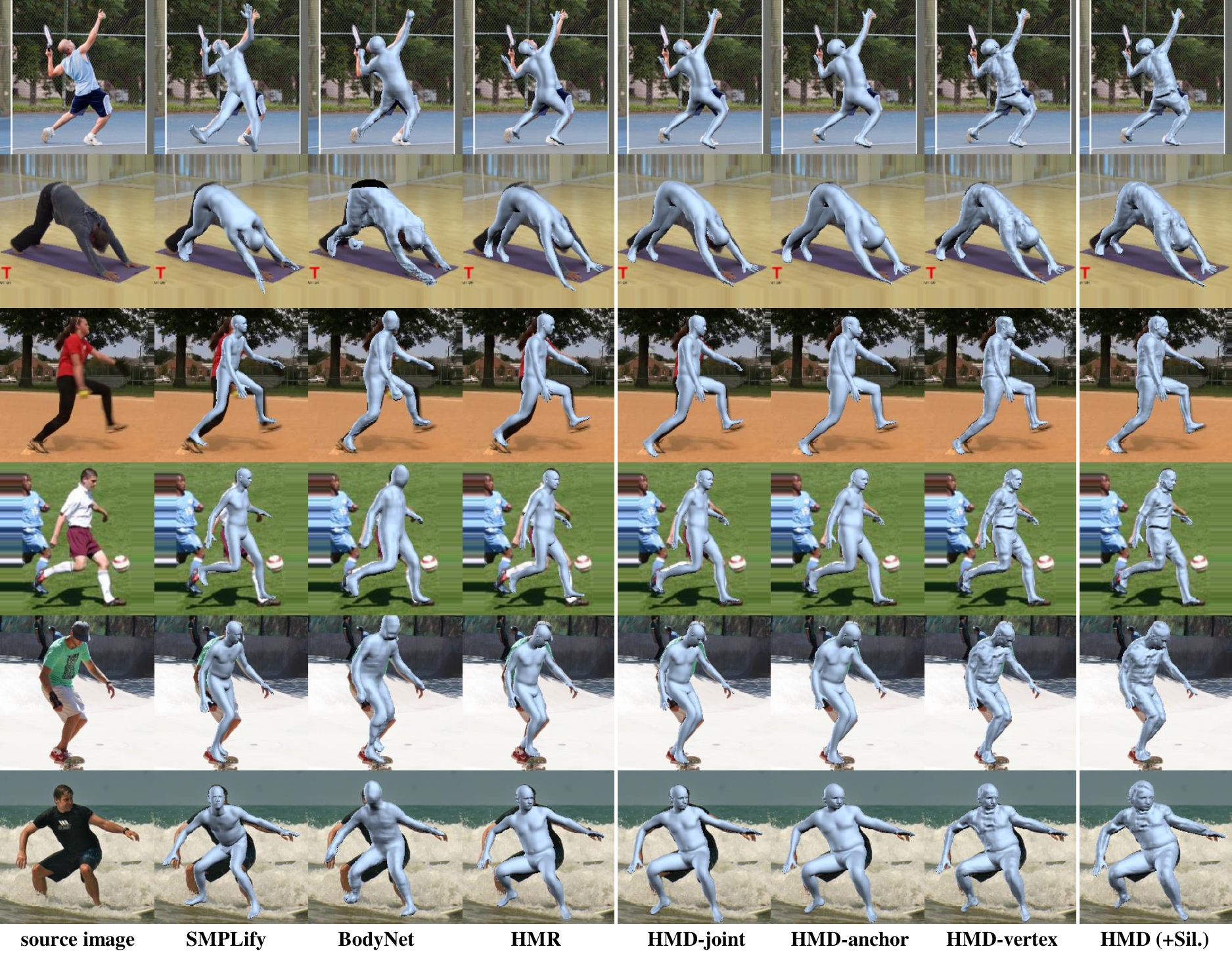}
\end{center}
   \caption{We compare our method on 3D human model reconstruction with previous approaches. The results of our method in different stages are shown: joint deformed, anchor deformed and vertex deformed (final result). Comparing to other methods, our method recovers more accurate joints and the body with shape details. The human body shape fits better to the input image, especially in body limbs. The rightmost column shows we can get more accurate recovered shapes when ground truth human silhouette is enforced (labeled as \textit{HMD (+Sil.)}). Note that the images are cropped for the compact layout.}
\label{fig:compare}
\end{figure*}

\begin{figure*}[t]
\begin{center}
   \includegraphics[width=1.0\linewidth]{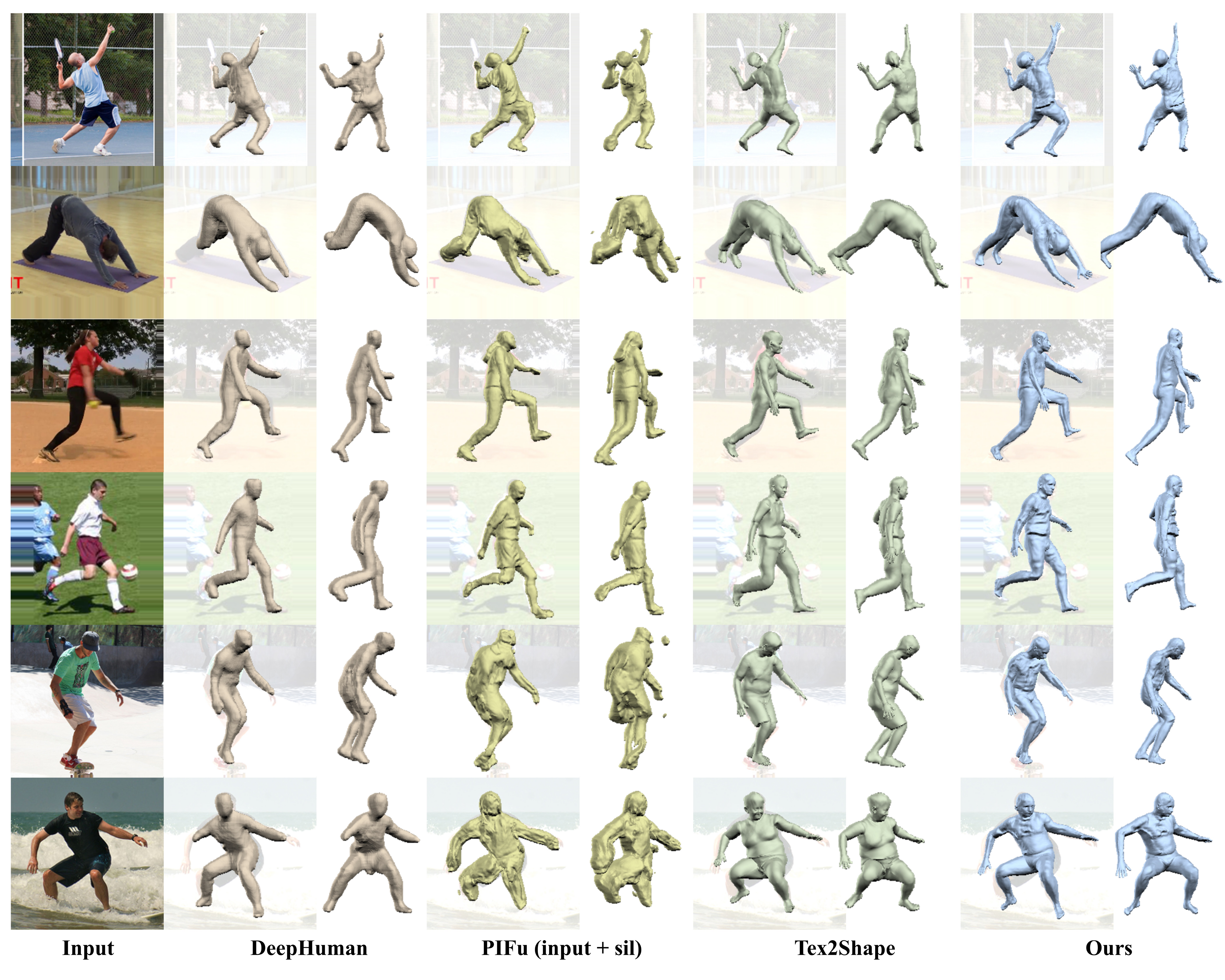}
\end{center}
   \caption{We compare our method with DeepHuman\cite{zheng2019deephuman}, PIFu\cite{saito2019pifu}, and Tex2Shape\cite{alldieck2019tex2shape}. These three methods are all trained using ground-truth 3D human shapes. It is worth noting that the input image to PIFu has been segmented using the ground-truth silhouette, while the inputs of the other methods are original images.}
\label{fig:compare_new}
\end{figure*}

\begin{figure*}[t]
\begin{center}
   \includegraphics[width=1.0\linewidth]{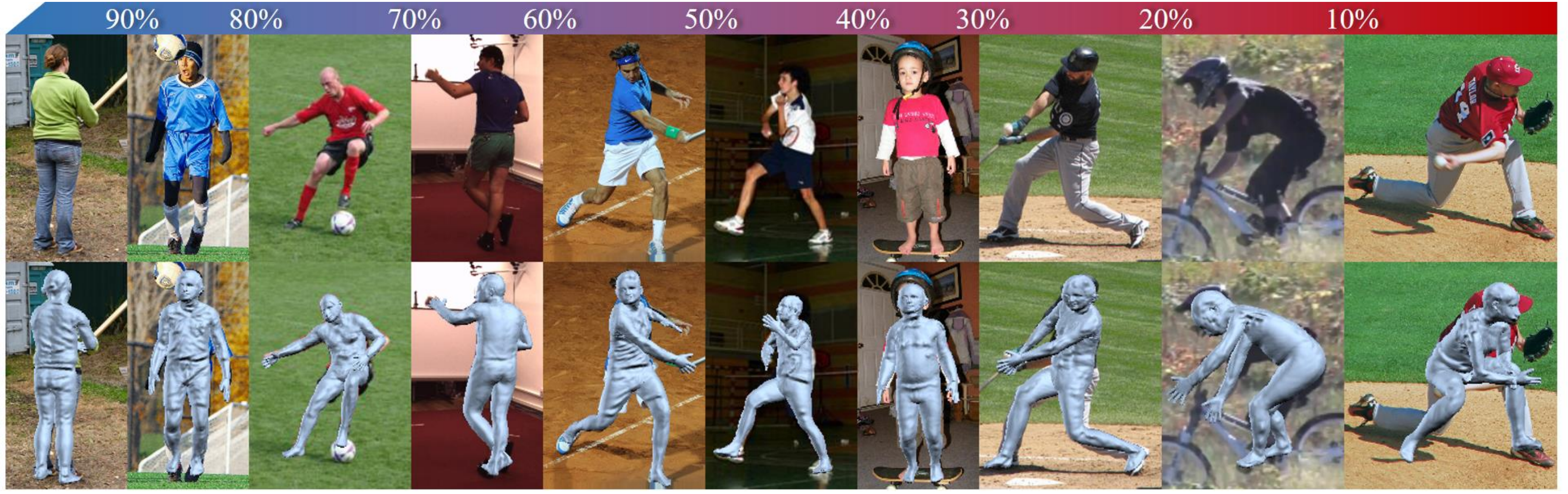}
\end{center}
   \caption{The results selected according to the percentage of the silhouette IoU.}
\label{fig:perc}
\end{figure*}

\begin{figure*}[t]
\begin{center}
   \includegraphics[width=1.0\linewidth]{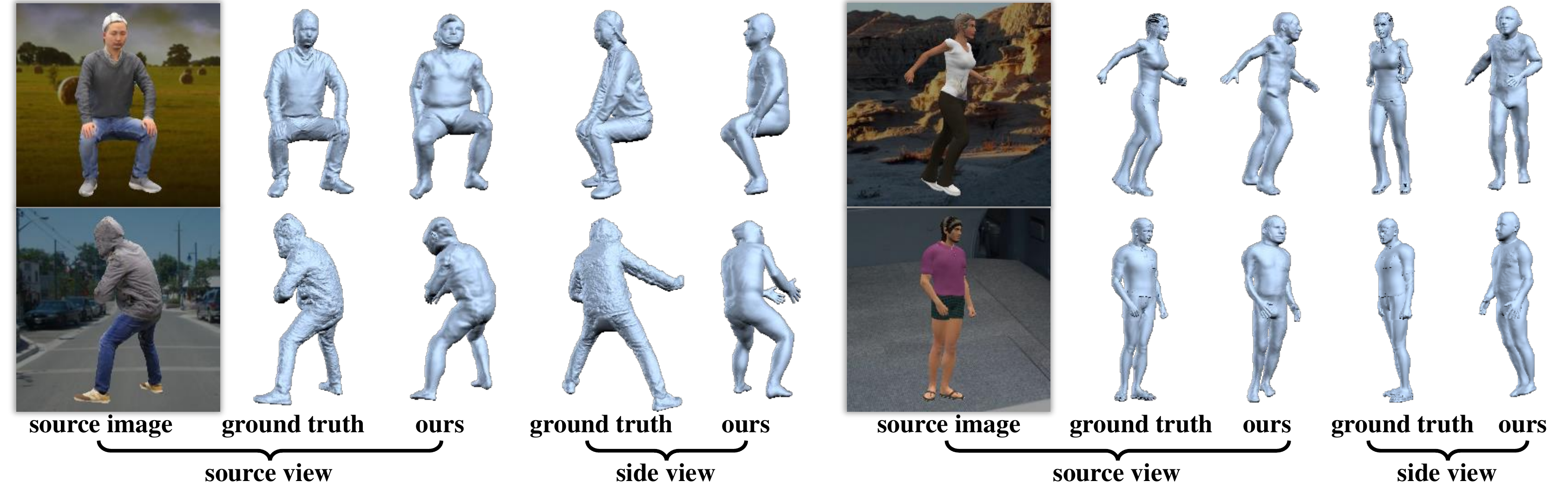}
\end{center}
   \caption{We show some recovered meshes and the ground truth meshes on the RECON (left) and SYN dataset (right).  The meshes are rendered in the side view by rotating the mesh $90^{\circ}$ around the vertical axis.}
\label{fig:real_syn}
\end{figure*}

\begin{figure*}[t]
\begin{center}
   \includegraphics[width=1.0\linewidth]{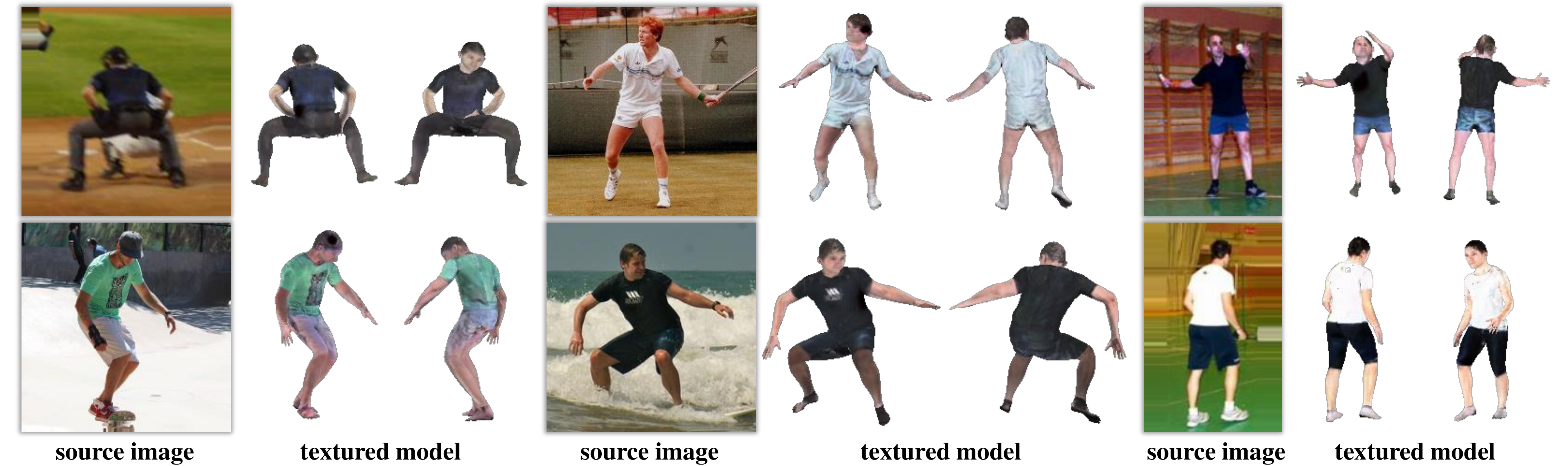}
\end{center}
   \caption{We show some texture synthesis results.  For each group, we show the source image and render the recovered mesh with predicted texture in the front view and back view.  Though there is slight color distortion comparing to the source image, our method is able to predict plausible texture, and can even hallucinate the completely invisible face from the back. }
\label{fig:result_texture}
\end{figure*}

\begin{figure}[t]
\begin{center}
   \includegraphics[width=1.0\linewidth]{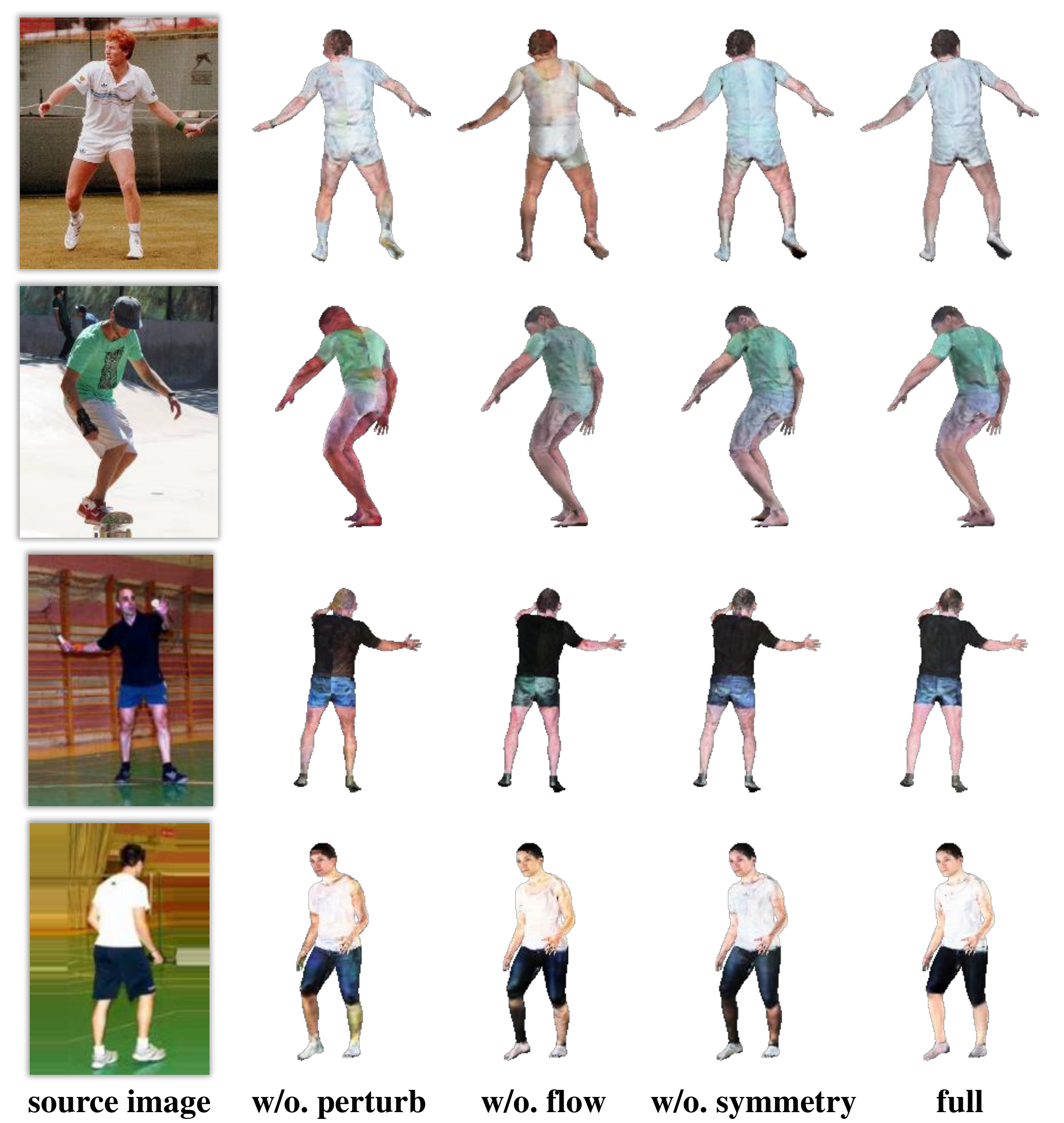}
\end{center}
   \caption{Qualitative ablation comparison of texture synthesis.}
\label{fig:ablation_texture}
\end{figure}

\begin{table*}[]
	\centering
\begin{threeparttable}
    \caption{Quantitative evaluation}
    \centering
    \begin{tabular}{lcccccccc}
    \hline
    & \multicolumn{2}{c}{-----WILD dataset-----} & \multicolumn{3}{c}{ ------------RECON dataset------------} & \multicolumn{3}{c}{ --------------SYN dataset--------------} \\
    method & sil IoU  & 2D joint err  & sil IoU  & 3D err full*  & 3D err visible*  & sil IoU  & 3d err full*  & 3d err visible*  \\ \hline
    SMPLify\cite{ECCV2016Bogo}      & 66.3   & 10.19   & 73.9   & 52.84   & 51.69   & 71.0   & 62.31   & 60.67   \\
    BodyNet\cite{ECCV2018Varol}     & 68.6   & ---     & 72.5   & 43.75   & 40.05   & 70.0   & 54.41   & 46.55   \\
    HMR\cite{CVPR2018Kanazawa}      & 67.6   & 9.90    & 74.3   & 51.74   & 42.05   & 71.7   & 53.03   & 47.75   \\
    HMD - joint                    & 70.7   & 8.81    & 78.0   & 51.08   & 41.42   & 75.9   & 49.25   & 45.70   \\
    HMD - anchor                   & 76.5   & 8.82    & 85.0   & 44.60   & 39.73   & 79.6   & 47.18   & 44.62   \\
    HMD - vertex                  & ---   & ---    & ---   & 44.10   & 41.76   & ---   & 44.75   & 41.90   \\ \hline
    HMD(+sil) - joint                 & 73.0   & 8.31    & 79.2   & 50.49   & 40.88   & 77.7   & 48.41   & 45.16   \\
    HMD(+sil) - anchor                & \textbf{82.4}   & \textbf{8.22}    & \textbf{88.3}   & 43.50   & \textbf{38.63}   & \textbf{85.7}   & 44.59   & 42.68   \\
    HMD(+sil) - vertex               & ---   & ---    & ---   & \textbf{43.22}   & 40.98   & ---   & \textbf{41.48}   & \textbf{39.11}   \\ \hline
    \end{tabular}
    \begin{tablenotes}
      \small
      \item * `full' means the full body shape is used for error estimation, and `vis' means only the visible part concerning the input image viewpoint is used for error estimation.
      \item The statistic unit of 3D error is millimeter; the 2D joint error is measured by pixel.  The methods beyond the cutting line use only RGB image as input, while the methods under the cutting line use `RGB + silhouette' as input.  Some statistic is blank: the joint position cannot be derived directly from the mesh produced by BodyNet;  The sil IoU and 2D joint error after vertex deformation stay the same as anchor deformed results, as the vertex deformation is only along the Z-axis, which is vertical to the silhouette in the image plane.
    \end{tablenotes}
\label{tab:quantitative}
\end{threeparttable}
\end{table*}

\subsection{Datasets}
\label{sec:dataset}
We have assembled three datasets for the experiment: the WILD dataset which has a large number of images with sparse 2D joints and segmentation annotated, and two other small datasets for evaluation in 3D metrics.

\textbf{WILD Dataset.}  We assemble a quite large dataset for training and testing by extracting from 5 human datasets including MPII human pose database (MPII)~\cite{CVPR2014Mykhaylo}, Common Objects in Context dataset (COCO)~\cite{ECCV2014Lin}, Human3.6M dataset (H36M)~\cite{TPAMI2014Ionescu, ICCV2011Ionescu}, Leeds Sports Pose dataset (LSP)~\cite{BMVC2010Johnson} and its extension dataset (LSPET)~\cite{CVPR2011Johnson}. As most of the images are captured in an uncontrolled setup, we call it the WILD dataset. The Unite the People (UP) dataset~\cite{CVPR2017Lassner} provides ground truth silhouettes for the images in LSP, LSPET, and MPII datasets.  As we focus on human shape recovery of the whole body, images with partial human bodies are removed based on the following two rules:

\begin{itemize}
\item All joints exist in the images.
\item All joints are inside the body silhouette.
\end{itemize}

For COCO and H36M dataset, we further filter out the data with low-quality silhouettes.
We separate the training and testing data according to the rules of each dataset.  The numbers of the data we use are listed in Table \ref{tab:datasets}.  

\begin{table}[]
\centering
\caption{WILD dataset components}
\begin{tabular}{ccccccc}
\hline
data source   & LSP & LSPET & MPII & COCO & H36M  \\ \hline
train num & 987 & 5376  & 8035 & 4004 & 5747  \\
test num  & 703 & 0     & 1996 & 606  & 1320  \\ \hline
\end{tabular}
\label{tab:datasets}
\end{table}

The main drawback of WILD dataset is the lack of 3D ground truth shape.  Though the UP dataset provides the fitted SMPL mesh for some data, the accuracy is uncertain.  To help evaluate the 3D accuracy, we make two other small datasets with ground truth shape:

\textbf{RECON Dataset}  
We reconstruct 25 human mesh models using the traditional multi-view 3D reconstruction methods~\cite{FTCGV2015Furukawa}. 
We render each model to 6 views and the views are randomly selected from 54 candidate views, of which the azimuth ranges from 0$^\circ$ to $340^\circ$ with intervals $20^\circ$, and the elevation ranges from $-10^\circ$ to $+10^\circ$ with intervals of $10^\circ$.  We use various scene images from the Places dataset\cite{TPAMI2017Zhou} as background. 

\textbf{SYN Dataset}  
We render 300 synthetic human mesh models in PVHM dataset~\cite{CVPR2018Zhu} following their rendering setup, with the random scene images from the Places dataset as background. The meshes of PVHM include the inner surface, which is a disturbance for surface accuracy estimation. To filter out the inner surface, we project the mesh to the viewpoints in 6 orthogonal directions and remove the faces which are invisible in all 6 viewpoints.

For RECON dataset and SYN dataset, the reconstructed 3D meshes are scaled so that the mean height of the human body is close to the general body height of the common adult. In this way, we could measure the 3D error in mm.

\begin{table}[]
\centering
\caption{Ablation experiments. In this table, (w) means the results when taking the window-cropped as input, and (f) means the results with the full image as input.}
\centering
\begin{tabular}{clcc}
\hline
num & method                 & sil IoU   & 2D joint err/ pixel  \\ \hline
1   & baseline(initial shape)      & 67.6\%      & 9.90          \\
2   & joint (f)                    & 68.3\%      & 9.85          \\
3   & joint (w)                    & 70.7\%      & 8.81          \\
4   & anchor (f)                   & 70.1\%      & 9.89          \\
5   & anchor (w)                   & 71.3\%      & 9.75          \\
6   & joint (w) + anchor (w)       & 76.5\%      & 8.82          \\ \hline
\end{tabular}
\label{tab:ablation}
\end{table}

\begin{table}[]
\centering
\caption{Ablation study of texture synthesis.}
\begin{tabular}{cccc}
\hline
method                 & PSNR   & MAE   & SSIM            \\ \hline
w/o. perturb    & 17.62  & 0.1710 & 0.5919         \\
w/o. flow              & 19.48  & 0.1223 & 0.6699         \\
w/o. symmetry          & 19.58  & 0.1200 & 0.6780         \\
Full                   & \textbf{19.65}  & \textbf{0.1189} & \textbf{0.6829}         \\ \hline
\end{tabular}
\label{tab:ablation_texture}
\end{table}

\subsection{Performance Evaluations}  
We measure the accuracy of the recovered shape with several metrics (corresponding to the second row in Table~\ref{tab:quantitative}). 
For all test sets, we report the silhouette Intersection over Union (referred to as sil IoU), which measures the matching rate of the projected silhouette of the predicted 3D shape and the image silhouette.
For the WILD dataset, we measure the difference between the projected 2D joints of the predicted 3D shape and the annotated ground truth joints.  The joints of the mesh are extracted by computing the geometric center of the corresponding joint handle vertices.  For the RECON dataset and SYN dataset, we also report the Chamfer distance (referred to as 3D err), which is the average distance of vertices between the predicted mesh and the ground truth mesh. We find the closest vertices in the resulting mesh for each vertex in the ground truth mesh and compute the mean of their distances as the 3D error. 

The results selected based on the rank of silhouette IoU are shown in Figure \ref{fig:perc}. We could see in columns of the left side, the person with a simple pose like standing yields a pretty good fit.  As we go from left to right columns, sports in the images are getting more complicated and the corresponding human shape is harder to predict. And in the right side columns, our method fails to predict humans with accessories (helmet, gloves) and under extremely twisting poses.  In summary, the performance of our method is mostly affected by the complexity of the human pose and articulation.

\subsection{Staging Analysis}  
\label{sec:stage_exp}

We show the staging results in Figure \ref{fig:compare} (right four columns) and report the quantitative evaluation of each stage in Table \ref{tab:quantitative}.  The results in different phases are named as HMD-joint, HMD-anchor, and HMD-vertex (final result).  We can see that the shape has got refined stage by stage.  In the joint deformation phase, the joint correction takes effect to correct the displacement of joints. In the anchor deformation phase, silhouette supervision plays a key role in fitting the human shape. In the vertex deformation stage, the shape details are recovered to produce a visually plausible result.

\textbf{Ablation study.}  We report the result of the ablation experiment in Table \ref{tab:ablation}, where (w) means the window-cropped input, and (f) means the full image input.  The evaluation is based on the WILD dataset, which is referred to in Section \ref{sec:dataset}.  We demonstrate two following statements: (1) By comparing the performance between full image input (No. 2 and 4) and window-crop image input (No. 3 and 5) in the table, we could see that the window-crop input predicts much higher silhouette IoU and lower joint error comparing to full image input, while the model size of the window-crop network is only $41\%$ of the full image network.  
The reason why it has got a better result is that the window-crop network inherently focuses on the handle as the input center, so the problem turns to predict the local fit for each handle, which is easier to learn.
(2) By comparing the performance between the integration of `joint + anchor' deformation (No. 6) and only anchor or joint deformation (No. 3 and 5), we find that the combination achieves the best performance, and shows larger improvement than the pure anchor deformation. 

To further validate the photometric loss as we train the ShadingNet, we also did the ablation study for the photometric loss term and computed the error of the refined depth on our captured depth dataset as described in Section 3.3. In total, we have 2272 frames and we randomly select 1818 frames for training and other 455 frames for testing. As shown in the table below, the surface error decreased after incorporating the photometric loss.

\begin{table}[htbp]
\centering
\caption{Ablation study of photometric loss.}
	{
		\begin{tabular}{l c }
			\hline
			\multicolumn{1}{c}{method}& \multicolumn{1}{c}{error(mm)}  \\ \hline
			Without Photometric loss  & 1.346\\
			With Photometric loss &  1.129  \\ \hline

		\end{tabular}
	}
	\label{tab:photoloss}
\end{table}

\textbf{Prediction with the silhouette.} By default our method takes the RGB image as input, and it also can use additional silhouettes as input.  The method using additional silhouette shares the same framework with the default setting, and the difference is explained in Section \ref{sec:deformation}.  We show the qualitative comparison result in the last column in Figure \ref{fig:compare} and the quantitative result in the last three rows in Table \ref{tab:quantitative}.  As expected, the prediction with silhouette produces better results in all metrics. 

\subsection{Comparison with Other Methods}
We compare our method with other methods with qualitative results as shown in Figure \ref{fig:compare} and quantitative results in Table \ref{tab:quantitative}.  We use the trained model of BodyNet and HMR provided by the authors.  As BodyNet requires 3D shapes for training, they don't use COCO and H36M datasets.   To be fair, the evaluation on the WILD datasets only uses the data from LSP, LSPET, and MPII, which are the intersection of datasets used in all estimated methods. 
Comparing to SMPL based methods (SMPLIify and HMR), our method has got the best performance in all metrics on all three datasets. As compared with BodyNet, a volumetric-based prediction method, we have got comparable scores in 3D error on RECON dataset. The reason is that the BodyNet produces more conservative shapes instead of focusing on the recovery of a complete human model. In some cases, the body limbs have not got reconstructed by the BodyNet when they are not visible from the image, while we always have the complete body recovered even though some parts of limbs haven't appeared in the image. This makes it easy to have a better registration to the ground-truth mesh resulting in smaller 3D error. However, their scores on SYN datasets are lower than the other two datasets, since the human subjects from the SYN dataset generally have slim body shapes in which case the BodyNet results are degraded.

\subsection{3D Error Analysis}

Figure \ref{fig:real_syn} shows our recovered 3D model on the RECON and SYN datasets together with the ground truth mesh. We show that the inherent pose and shape ambiguities cannot be resolved with the image from a single viewpoint. As we can see in Figure \ref{fig:real_syn}, the human shapes seen from the side view are quite different from the ground truth model even though they could fit closely to the input image. The estimated depth cue from a single image is sometimes ambiguous for shape recovery.  This observation explains the reason why the improvement of our method in 2D metrics is relatively larger than the improvement in 3D metrics. 

We also evaluate MPJPE after rigid alignment as `3D joint error' (defined as `Reconst. Error' in HMR\cite{CVPR2018Kanazawa}) on H36M dataset.  To derive joints from our models, we use the unified joints definition defined in Section 3.1.  The results are reported in Table \ref{tab:error_h36m}.  We can see that:
(1) Our method is superior in both 2D and 3D joint error comparing to HMR.  The improvement of 3D joint error is slightly smaller than that of 2D joint error because there are few changes in the depth direction.
(2) The 2D error of H36M dataset is generally smaller than that of WILD dataset, because the poses in H36M are relatively simpler than the other data in WILD dataset.
(3) Adding silhouette as input leads to smaller improvement in H36M dataset than in WILD dataset, because the background in H36M images is pure colored, so there is less distraction from varying backgrounds.  

\begin{table}[htbp]
	\caption{Joint error evaluation on H36M dataset. HMD-j means the result after joint deforming stage, and HMD-a means the result after joint and anchor deforming stage. (s) means using the ground-truth silhouette as input. }  
	\centering  
	\label{tab:error_h36m}
	\begin{tabular}{c|ccccc} 
		\hline  
		method        & HMR    & HMD-j    & HMD-a    & HMD(s)-j    & HMD(s)-a \\  \hline
		3D err (mm)   & 56.5   & 49.6     & 50.9     & 49.2        & 50.4      \\
		2D err (mm)   & 5.42   & 3.96     & 4.19     & 3.77        & 4.11      \\
		\hline
	\end{tabular}
\end{table}

\subsection{Texture Synthesis}
The texture synthesis results are shown in Figure~\ref{fig:result_texture}, from which we can see that our method can synthesize plausible texture for the invisible part and even for the human face on the reverse side. 

\noindent\textbf{Ablation Study.} We perform the ablation study to demonstrate the effectiveness of each component in our texture synthesis framework. We randomly choose 836 texture maps in SURREAL dataset as training data, leaving 93 texture maps for testing. The pose parameters in UP dataset are divided into $90\%$ for training and $10\%$ for testing. Then we synthesize the training and testing set as explained in Section \ref{ssec:Texture_trainging_data}. The evaluation results are shown in Table~\ref{tab:ablation_texture}, where the explanation of each test are: 

\begin{itemize}
\item Without perturb. We train our network on a synthetic dataset without adding random noise. Specifically, we detect visible faces on the mesh according to the ground truth SMPL and viewpoint parameters. Then we only keep visible parts in the texture as our partial texture.

\item Without flow. This model has the same architecture as our refinement network. We take the partial texture directly as input and train the network to hallucinate the missing region.

\item Without symmetry. This model is identical to our complete method except that the input to the refinement network doesn't contain the symmetrically flipped texture.

\item Full. This is the complete architecture we described in the texture synthesis section.
\end{itemize}

Table \ref{tab:ablation_texture} and Figure \ref{fig:ablation_texture} show the quantitative and qualitative results of the ablation study.   From the qualitative comparison, we can see that the results without perturbing contain textures from the background, which means the model cannot tolerant the interfering texture caused by inaccurate shape prediction.  The results without flow contain obvious color distortion, which is a common problem for image generation networks.  The results without symmetry also show color distortion, and the seam of the UV texture map in the back is inconsistent.  The results of full architecture are visually most plausible in the ablation experiments.  In the quantitative comparison, we can see that the network trained on the dataset without noise predicts worse results on the testing set, as the recovered mesh cannot be perfectly aligned to the image. Our full architecture scores the highest in all metrics.

\begin{figure*}
\begin{center}
   \includegraphics[width=1.0\linewidth]{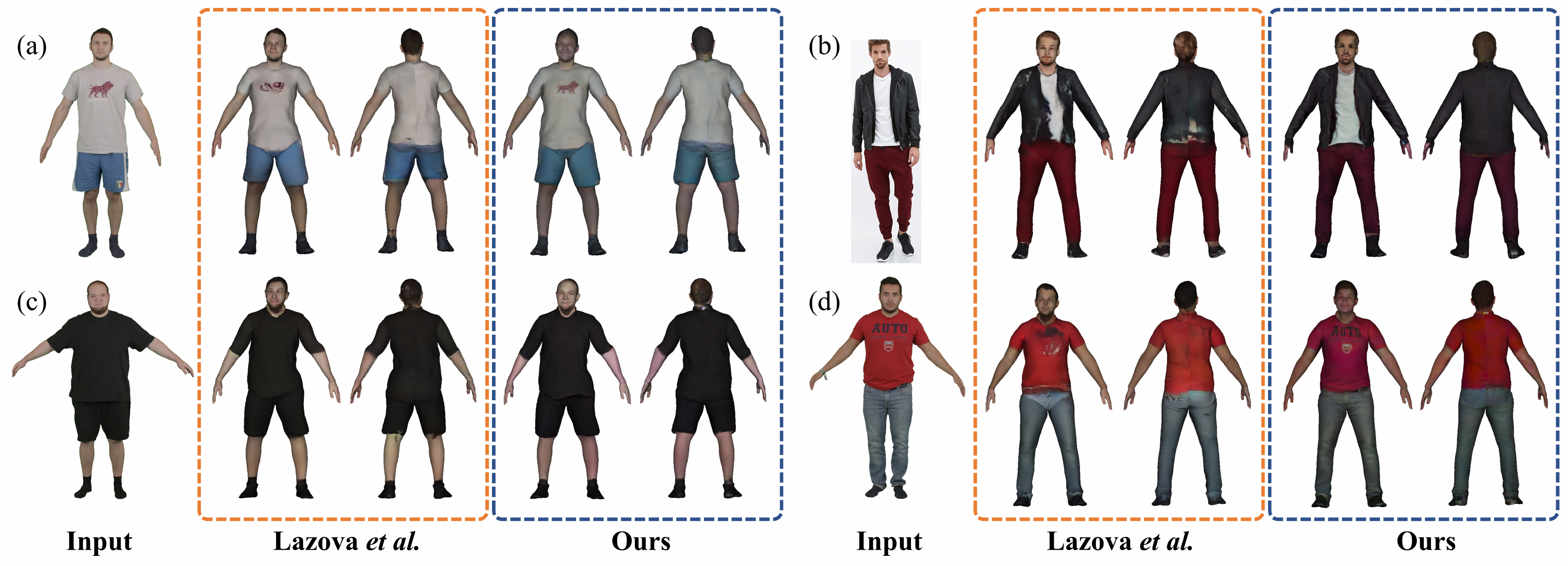}
\end{center}
   \caption{Comparison between our texture completion and Lazova \textit{et al.}'s work~\cite{lazova2019360}.}
\label{fig:tex_compare}
\end{figure*}

\noindent\textbf{Comparison.} We compare our texture completion method with Lazova \textit{et al.}'s work~\cite{lazova2019360} in Figure \ref{fig:tex_compare}. The source images, models, and the results are provided by the authors.  To make the comparison clear, the predicted textures are mapped to the same canonical A-pose meshes. Lazova \textit{et al.}'s work~\cite{lazova2019360} uses the correspondence estimated by DensePose\cite{CVPR2018Guler} to generate the input partial texture, while our method uses the projective texture from the recovered human shape as input.  By comparison, we can see that Lazova \textit{et al.}'s work generates slightly better facial texture, as DensePose aligns better texture alignment in the facial region.  Our method can preserve more texture details in the main body. For example, the logo on the T-shirt in (a) and (d), and the coat edges in (b) are preserved by our method but are blurred in Lazova \textit{et al.}'s result.  Besides, Lazova \textit{et al.}'s inpainting model also introduces some artifacts and blurriness into the restored texture. Our method shows better texture quality in the back area due to the sophisticated two-stage flow-refine scheme. Our symmetry-input design alleviates the inconsistent seam in the back that Lazova \textit{et al.}'s work suffers, as shown in the back view of (b) and (d).

\subsection{Comparison with 3D-Supervised Methods}

\begin{table*}[]
\centering
\begin{threeparttable}
\caption{Quantitative comparison with 3D supervised methods.}
\centering
\begin{tabular}{lccccccccc}
\hline
                & \multicolumn{3}{c}{---------RECON dataset---------} & \multicolumn{3}{c}{---------SYN dataset---------} & \multicolumn{3}{c}{---------BUFF dataset---------} \\
method          & err full     & err visible     & sil IoU    & err full    & err visible    & sil IoU    & err full     & err visible    & sil IoU    \\ \hline
PIFu$^*$\cite{saito2019pifu}            & 35.67$^*$          & 32.77$^*$             & --$^*$        & 18.96$^*$         & 17.68$^*$            & --$^*$        & 26.65$^*$          & 27.33$^*$            & --$^*$        \\
PIFuHD\cite{saito2020pifuhd}          & 52.69        & 48.55           & 58.4       & 22.17       & 20.55          & 56.53      &\textbf{22.01}&\textbf{24.90}  &\textbf{87.8}\\
DeepHuman\cite{zheng2019deephuman}       &\textbf{38.58}&\textbf{37.26}   & 71.2       &\textbf{19.34}&\textbf{18.60} & 60.3       & 34.97        & 37.49          & 83.1       \\
Tex2Shape\cite{alldieck2019tex2shape}       & 44.7         & 44.39           & 61.2       & 26.86       & 27.82          & 54.8       & 58.80        & 61.03          & 55.3       \\
HMD             & 44.1         & 41.76           &\textbf{85.0}& 44.75      & 41.90          &\textbf{79.6}& 41.30       & 40.02          & 80.7       \\ \hline
\end{tabular}
    \begin{tablenotes}
      \small
      \item Please refer to Table~\ref{tab:quantitative} for the explanation of metrics.
      \item * PIFu requires an additional ground-truth mask as input, while all other methods take only source images as input.
      
    \end{tablenotes}
\label{tab:comp_3dmethods}
\end{threeparttable}
\end{table*}

In Figure \ref{fig:compare_new}, we show some sample results for qualitative comparison with DeepHuman\cite{zheng2019deephuman}, PIFu\cite{saito2019pifu} and Tex2Shape\cite{alldieck2019tex2shape}. 
It is worth noting that these three methods all take the ground-truth 3D model for training, and are published after our submission. Specifically, PIFu used the high-quality 3D models from RenderPeople\cite{renderpeople}; Tex2Shape used 3D models from RenderPeople\cite{renderpeople}, Twindom\cite{twindom}, and axyzdesign.com; DeepHuman used the 3D human models captured with Kinect and reconstructed using DoubleFusion\cite{2018DoubleFusion}. By contrast, our joint net and anchor net of our method was trained on the dataset collected from in-the-wild images without 3D ground-truth models as supervision, and the shading net of our method requires the wild images and a small number of depth maps for training. 

In Figure \ref{fig:compare_new}, we can see that the DeepHuman\cite{zheng2019deephuman} can recover the complete body in most cases, but it has hands/feet missing in some cases (line 1/2/6). Generally, the recovered mesh of DeepHuman lacks 3D geometric details. 
PIFu\cite{saito2019pifu} can recover detailed wrinkles in the front view, but its performance is poor in some complex poses, leading to abnormal shape (line 2/3/5/6), broken arms (line 1), and duplicated limbs (line 3).  This is caused by the limited diversity of poses in RenderPeople, most of which are upright-posed human models. 
Tex2Shape has a good ability to maintain the overall shape and could recover middle-scale garment shape. However, the reconstructed surface is over-smoothed and lacks geometric details.  
Our method is more stable for the in-the-wild images and is able to reconstruct correct human poses and shapes together with the detailed geometric structure aligned well with the input image.

We also conduct quantitative comparison with the 3D supervised methods mentioned above and report the surface and IoU error in Table~\ref{tab:comp_3dmethods}. We run the evaluation on RECON dataset, SYN dataset, and BUFF dataset\cite{Zhang_2017_CVPR}. The details of RECON and SYN datasets have been presented in Section~5.1. For the BUFF dataset, it consists of 100 pairs of models and images, with 5 subjects performing daily motions. We render images for the ground-truth 3D model in BUFF dataset following the same settings when generating the images in RECON dataset. PIFuHD and PIFu performed better in BUFF dataset where the subjects are in relatively simple standing poses but the results are worse in RECON dataset where complex poses like crouching and lunges are involved. PIFu needs to take an additional ground-truth mask as input at test run. On the other hand, DeepHuman adopted the well-fitted SMPL as input, which improved the robustness for complex poses, and thus achieved stable performance overall. These comparison methods achieved better results with smaller 3D surface error but they all rely on ground-truth 3D models for training while our method only uses 2D supervision, such as 2D joints. We got comparable results on RECON dataset containing humans under complex poses, and a higher score in IoU of the silhouette.

\subsection{Performance of 3D-supervised Training}
\label{sec:3dspv}
\begin{figure*}[t]
\begin{center}
   \includegraphics[width=1.0\linewidth]{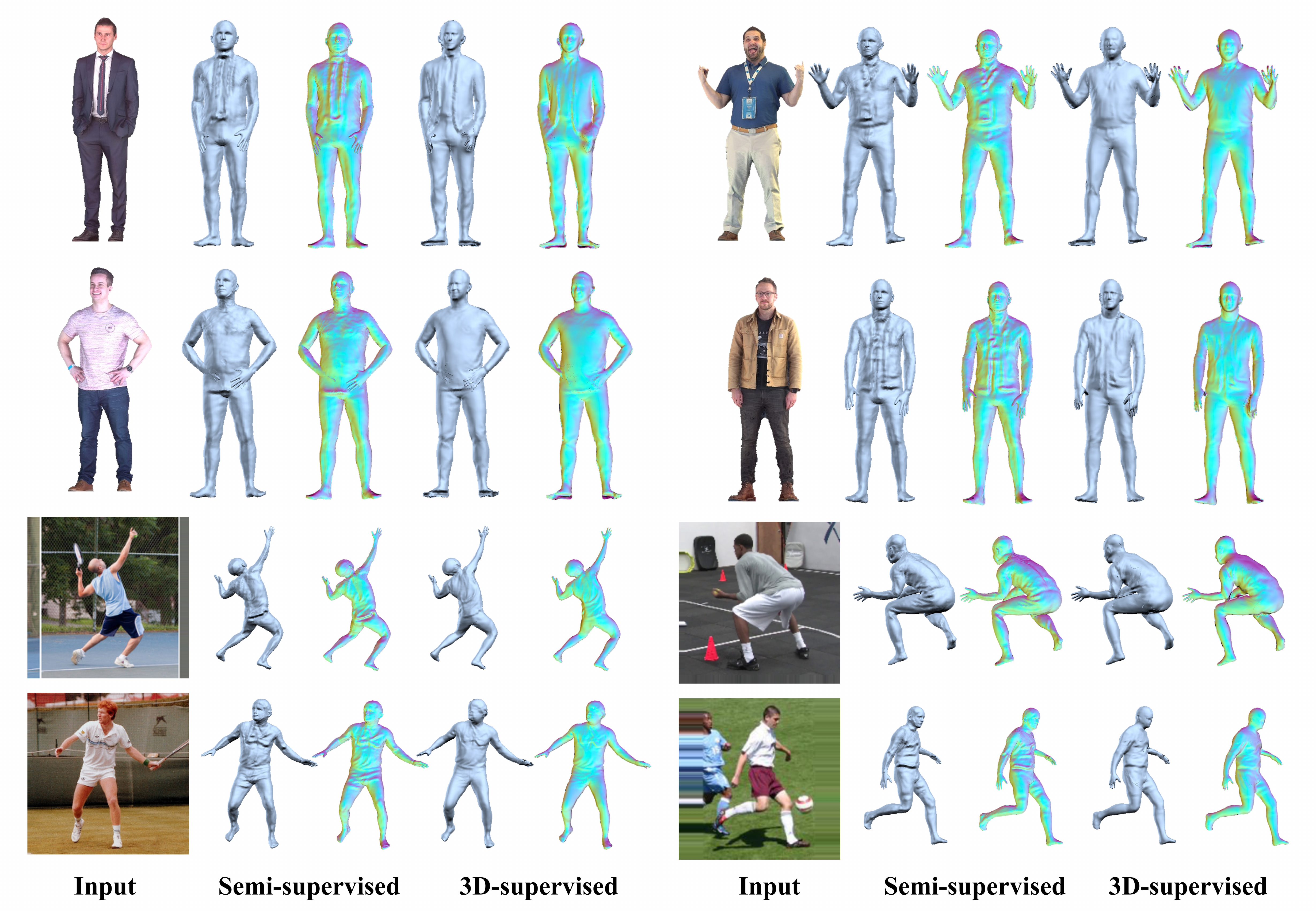}
\end{center}
   \caption{Comparison between semi-supervised results and 3D-supervised results on the Twindom dataset (upper two rows) and WILD dataset (lower two rows).  In each set of results, the left side is the rendered mesh model, and the right side is the rendered normal map.}
\label{fig:3d-spv}
\end{figure*}

The results of 3D-supervised training are shown in Figure \ref{fig:3d-spv}. By comparing the 3D-supervised and semi-supervised results, we can see both schemes can recover geometric details like the wrinkles and the belt, but the 3D-supervised results are closer to the real human 3D shape with fewer texture-copying effects. For example, for the second person in the second column of the figure, the semi-supervised method has taken the texture of the shirt as the geometric surface details, and thus created the wrong wrinkles. Besides, the semi-supervised method tends to yield excessive shape change.  By contrast, 3D-supervised schemes recover stable 3D detailed shape, and we believe this is because the Twindom dataset provides abundant shape and appearance together with the 3D human shape. But the limitation of the 3D-supervised scheme is that it relies on an expensive commercial high-quality 3D human dataset. 

\subsection{View Synthesis}

\begin{figure}[t]
\begin{center}
   \includegraphics[width=1.0\linewidth]{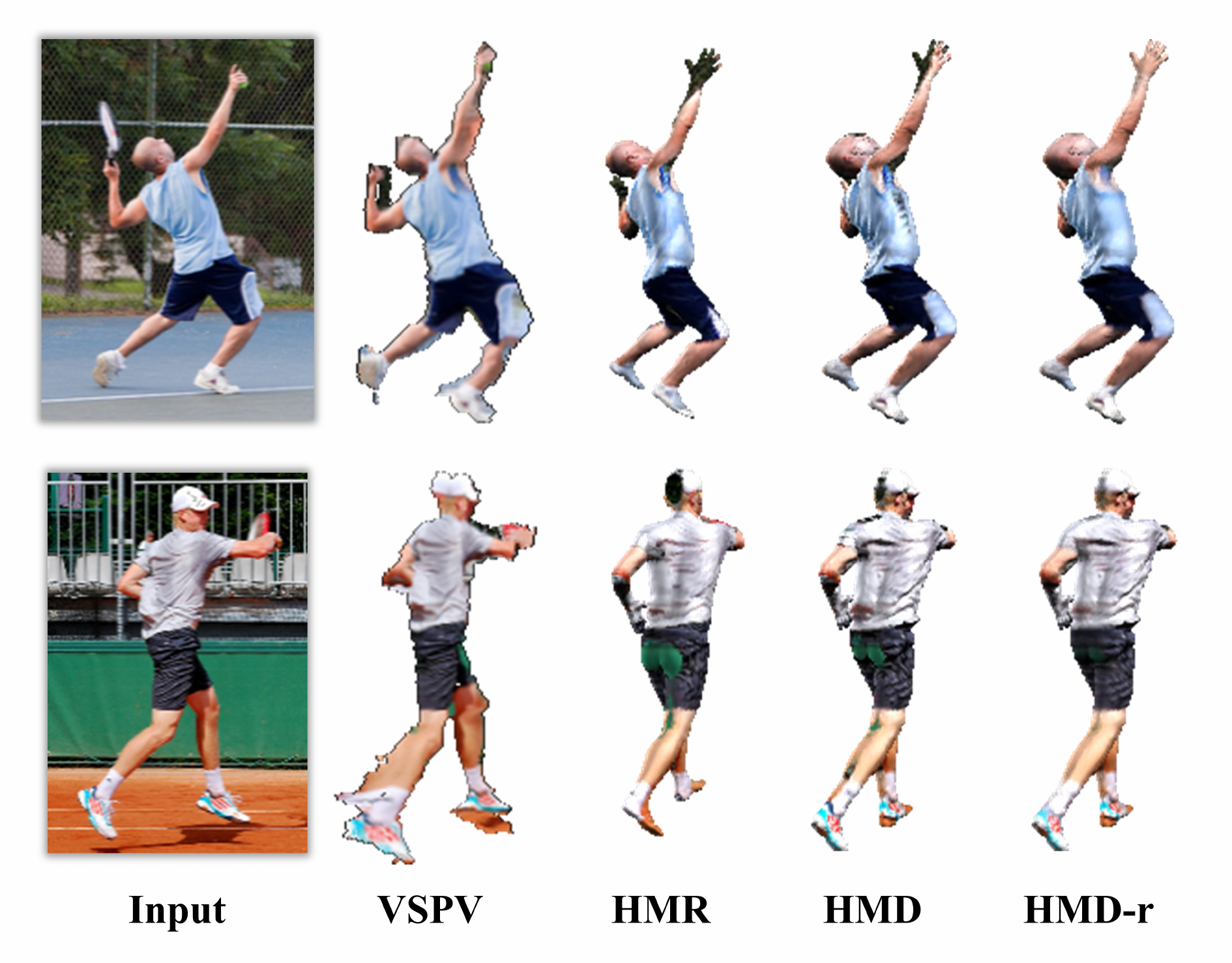}
\end{center}
   \caption{View synthesis results using our predicted model and other methods.}
\label{fig:view_synthesis}
\end{figure}

In Figure~\ref{fig:view_synthesis}, we use the predicted model by our method and HMR\cite{CVPR2018Kanazawa} for view synthesis task. To this end, we first assign the color of the pixels in the source image to the vertices of the predicted mesh, then render the model in the novel view to generate the result images. To further refine the synthesis result, we expand the texture in the foreground part to avoid artifacts in the boundary region, which is labeled as 'HMD-r' in the figure.  As the predicted HMR model couldn't fit well to the source image, the artifacts exist in the boundary region. By contrast, we get much better results with visually appealing synthesis using our texture completion method. We also compare our results with VSPV\cite{CVPR2018Zhu}, which is an image-to-image synthesis method. We find that VSPV fails to track the right pose when the pose is complex, and the rotation of viewpoint is large (60 degrees in the figure). By contrast, our method synthesizes more plausible images.

\section{Conclusion}

In this paper, we have proposed a novel approach to reconstruct the detailed human avatar from a single image, which is the 3D human shape with texture. Starting from an SMPL model based human recovery method, we introduce free-form deformations to refine the body shapes with a project-predict-deform strategy.  A hierarchical framework has been proposed for restoring more accurate and detailed human bodies under the supervision of joints, silhouettes, and shading information.  Furthermore, we use the neural network to synthesize the complete texture for the recovered 3D mesh using the single image.  We have performed extensive comparisons with state-of-the-art shape recovering methods and demonstrated significant improvements in both quantitative and qualitative assessments.   Our method also generates plausible texture for the recovered 3D model, and can even hallucinate invisible face from the back.

The limitation of our work is that the pose ambiguities are not solved, and there are still large errors in predicted body meshes especially in depth direction.  The results for the human in relatively unusual poses are sometimes not good.  For the texture completion part, due to the limitation of the training dataset, the predicted texture is inclined to the appearance in SURREAL dataset.

\section*{Acknowledgements}
This work was supported by the NSFC grant 62025108, 62001213, 61627804, USDA grant 2018-67021-27416, and NSERC Discovery Grant.

\ifCLASSOPTIONcaptionsoff
\newpage
\fi



%
\bibliographystyle{ieee}
\bibliography{mybib}
%

\begin{IEEEbiography}
	[{\includegraphics[width=1in,height=1.25in,clip,keepaspectratio]{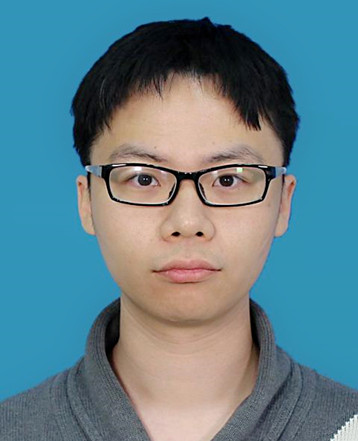}}]
	{Hao Zhu} received the Ph.D. and B.S. degree from the School of Electronic Science and Engineering, Nanjing University, Nanjing, China. He was a visiting scholar in University of Kentucky.  He is currently an associate researcher in Nanjing University.  His current research interests include computer vision and deep learning, especially 3D reconstruction and 3D vision. 
\end{IEEEbiography}

\begin{IEEEbiography}
	[{\includegraphics[width=1in,height=1.25in,clip,keepaspectratio]{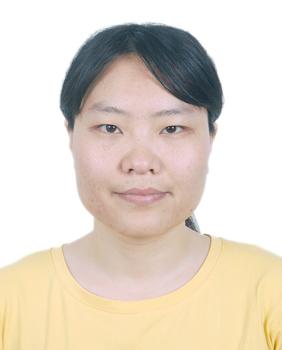}}]
	{Xinxin Zuo} received the M.E. degree from Northwestern Polytechnical University and Ph.D. degree from the University of Kentucky. She is currently a Postdoctoral Fellow at University of Alberta. Her research interests include computer vision and graphics, especially on 3D reconstruction and human modeling. 
\end{IEEEbiography}

\begin{IEEEbiography}
	[{\includegraphics[width=1in,height=1.25in,clip,keepaspectratio]{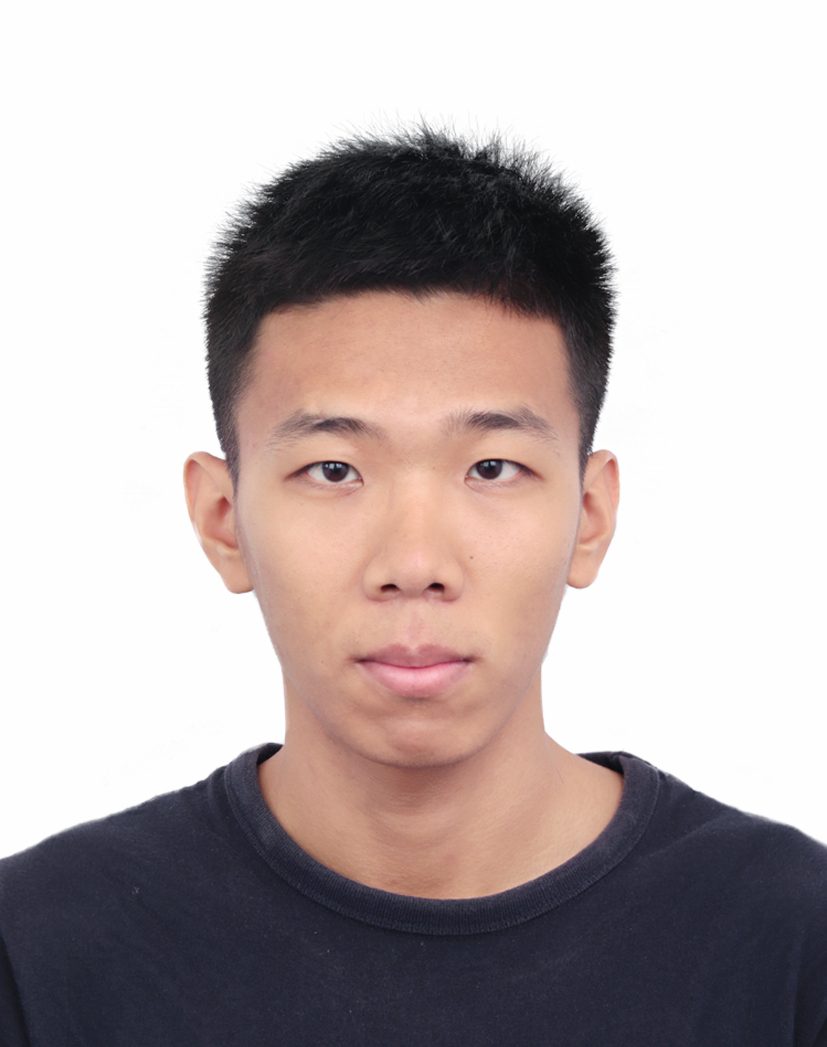}}]
	{Haotian Yang } received the B.S. degree from Nanjing University, Nanjing, China, in 2018, where he is currently pursuing the M.S. degree with the School of Electronic Science and Engineering.
	His current research interests include computer vision and computer graphics. 
\end{IEEEbiography}

\begin{IEEEbiography}
	[{\includegraphics[width=1in,height=1.25in,clip,keepaspectratio]{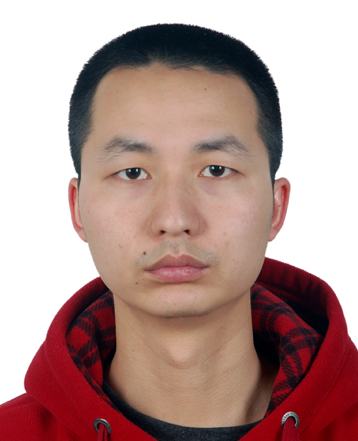}}]
	{Sen Wang} received the B.E. degree and Ph.D. degree from Northwestern Polytechnical University. From 2015 to 2016, he was a Visiting Ph.D. Student at the University of Kentucky. He is currently a Postdoctoral Fellow at University of Alberta. His research interests include computer vision and
robotics.
\end{IEEEbiography}

\begin{IEEEbiography}
	[{\includegraphics[width=1in,height=1.25in,clip,keepaspectratio]{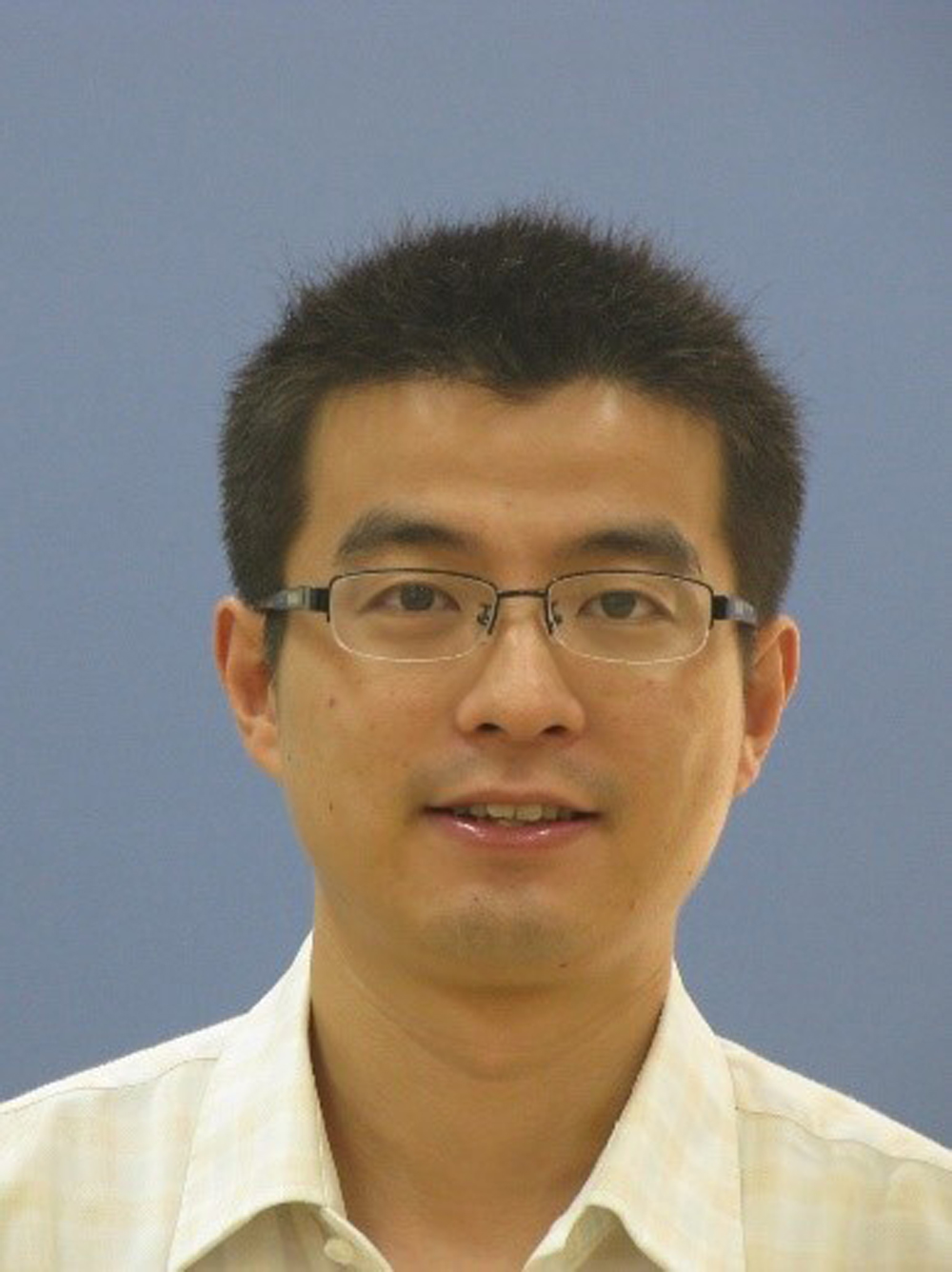}}]
	{Xun Cao} received the B.S. degree from Nanjing University, Nanjing, China, in 2006, and the Ph.D. degree from the Department of Automation, Tsinghua University, Beijing, China, in 2012. He held visiting positions with Philips Research, Aachen, Germany, in 2008, and Microsoft Research Asia, Beijing, from 2009 to 2010. He was a Visiting Scholar with the University of Texas at Austin, Austin, TX, USA, from 2010 to 2011. He is currently a Professor with the School of Electronic Science and Engineering, Nanjing University. His current research interests include computational photography and image-based modeling and rendering. 
\end{IEEEbiography}


\begin{IEEEbiography}
	[{\includegraphics[width=1in,height=1.25in,clip,keepaspectratio]{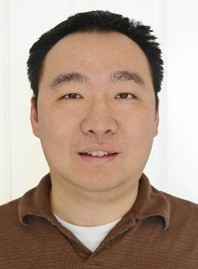}}]{Ruigang Yang}
	is currently a full professor of Computer Science at the University of Kentucky. He obtained his PhD degree from University of North Carolina at Chapel Hill and his MS degree from Columbia University. His research interests span over computer graphics and computer vision, in particular in 3D reconstruction and 3D data analysis. He has published over 100 papers, which, according to Google Scholar, has received 
more than 10000 citations with an h-index of 52 (as of 2019). He has received a number of awards, including US NSF Career award in 2004 and the Dean’s Research Award at the University of Kentucky in 2013. He is a senior member of IEEE.
\end{IEEEbiography}




\end{document}